\documentclass[10pt,twocolumn,letterpaper]{article}

\usepackage{iccv}
\usepackage{times}
\usepackage{epsfig}
\usepackage{graphicx}
\usepackage{amsmath}
\usepackage{amssymb}

\usepackage[accsupp]{axessibility} 

\usepackage{booktabs}
\usepackage{subcaption}
\usepackage{adjustbox}

\usepackage[pagebackref=true,breaklinks=true,letterpaper=true,colorlinks,bookmarks=false]{hyperref}

\newcommand{\T}{\mathcal{T}}
\newcommand{\W}{\mathcal{W}}
\newcommand{\loss}{\mathcal{L}} 
\newcommand{\Q}{\mathcal{Q}}

\newcommand{\deva}[1]{}
\newcommand{\chittesh}[1]{}
\newcommand{\nicolas}[1]{}
\newcommand{\martin}[1]{}

\newcommand{\ignore}[1]{}

\newif\ifappendix
\newif\ifstandalonesupplement
\newif\ifmuteappendixcite
\newif\ifunlimited

\iccvfinalcopy 

\appendixtrue 

\unlimitedtrue 

\def\iccvPaperID{3521} 

\begin{document}

\ifstandalonesupplement
    \title{Supplementary Material for FOVEA: Foveated Image Magnification for Autonomous Navigation}
\else
    \title{FOVEA: Foveated Image Magnification for Autonomous Navigation}
\fi

\author{Chittesh Thavamani\textsuperscript{1}\thanks{ \  denotes equal contribution.} \qquad
Mengtian Li\textsuperscript{1}\footnotemark[1] \qquad
Nicolas Cebron\textsuperscript{2} \qquad
Deva Ramanan\textsuperscript{1,2} \\
\textsuperscript{1}Carnegie Mellon University \qquad \textsuperscript{2}Argo AI
}

\maketitle


\ifstandalonesupplement
    \renewcommand{\thesection}{\Alph{section}}
    \renewcommand{\thefigure}{\Alph{figure}}
    \renewcommand{\thetable}{\Alph{table}}
    \ifmuteappendixcite
\renewcommand{\cite}[1]{}
\fi

\ifstandalonesupplement

In the supplementary material, we provide a demo video (``\texttt{Demo (2min).mp4}'') and this document. The demo video explains our proposed method and contains qualitative visualization. In this document, we provide additional diagnostic experiments for ablation studies, detection-only streaming evaluation, additional visualization, and additional implementation details.

\fi

\section{Additional Diagnostic Experiments}
\label{app:diagnostics}

\subsection{The Role of Explicit Backward Label Mapping}
\label{sec:no-back-map}

Related work either focus on tasks with labels invariant to warping like image classification or gaze estimation \cite{jaderberg2015spatial,recasens2018learning} (discussed in 
\ifstandalonesupplement
    Sec~3.1),
\else
    Sec~\ref{sec:background}),
\fi
or expect an implicit backward mapping to be learned through black-box end-to-end training \cite{marin2019efficient}
(discussed in
\ifstandalonesupplement
    Sec~2).
\else
    Sec~\ref{sec:related}).
\fi
In this section, we suggest that the implicit backward label mapping approach is not feasible for object detection. To this end, we train and test our KDE methods minus any bounding box unwarping. Specifically, we no longer unwarp bounding boxes when computing loss during training and when outputting final detections during testing. Instead, we expect the model to output detections in the original image space.

Due to instability, additional measures are taken to make it end-to-end trainable. First, we train with a decreased learning rate of 1e-4. Second, we train with and without adding ground truth bounding boxes to RoI proposals. The main KDE experiments do not add ground truth to RoI proposals, because there is no way of warping bounding boxes into the warped image space (the implementation of $\T$ does not exist). We additionally try setting this option here, because it would help the RoI head converge quicker, under the expectation that the RPN should output proposals in the original space. All other training settings are identical to the baseline setup (\ifstandalonesupplement Sec~4.1.1)\else Sec~\ref{Baseline and Setup})\fi.

Results are shown in Tab~\ref{tab:additional-diagnostics}. The overall AP is single-digit under all of these configurations, demonstrating the difficulty of implicitly learning the backward label mapping. This is likely due to the fact that our model is pretrained on COCO \cite{lin2014microsoft}, so it has learned to localize objects based on their exact locations in the image, and finetuning on Argoverse-HD is not enough to ``unlearn" this behavior and learn the backward label mapping. Another factor is that in the $S_I$ and $S_C$ cases, each image is warped differently, making the task of learning the backwards label mapping even more challenging. We suspect that training from scratch with a larger dataset like COCO and using the warp parameters (e.g. the saliency map) as input may produce better results. However, this only reinforces the appeal of our method due to ease of implementation and cross-warp generalizability (we can avoid having to train a new model for each warping mechanism).

\subsection{Sensitivity to Quality of Previous-Frame Detections}
\label{sec:perfect-saliency}

Two of our methods, $S_I$ and $S_C$ are dependent on the accuracy of the previous-frame detections. In this section, we analyze the sensitivity of such a dependency through a soft upper bound on $S_I$ and $S_C$, which is generated using the current frame's ground truth annotations in place of detections from the previous frame. This soft upper bound is a perfect saliency map, up to the amplitude and bandwidth hyperparameters. Note that this is only a change in the testing configuration.

We report results in Tab \ref{tab:additional-diagnostics}. We see a significant boost in accuracy in all cases. Notably, the finetuned KDE $S_I$ model at $0.5$x scale achieves an AP of $29.6$, outperforming the baseline's accuracy of $29.2$ at $0.75$x scale.

\subsection{Sensitivity to Inter-Frame Motion}
\label{sec:motion-sensitivity}

\begin{table*}[hbt!]
    \centering
    Argoverse-HD before finetuning\\
    \begin{adjustbox}{width=1.0\linewidth,center}
    \begin{tabular}{@{}lcccccccccccccc@{}}
    \toprule
    Method    & AP & AP$_{50}$ & AP$_{75}$ & AP$_{S}$ & AP$_{M}$ & AP$_{L}$ & person & mbike & tffclight & bike & bus & stop & car & truck \\
    \midrule
    \multicolumn{10}{@{}l}{\textbf{Main Results} (copied from the main text for comparison)} \\
    \ \ \ Baseline & 21.5	&35.8	&22.3	&2.8	&22.4	&\textbf{50.6}	&20.8	&9.1	&13.9	&7.1	&48.0	&16.1	&37.2	&20.2 \\
    \ \ \ KDE ($S_D$) & 23.3	&40.0	&22.9	&5.4	&25.5	&48.9	&20.9	&13.7&	12.2	&9.3&	\textbf{50.6}&	20.1&	40.0&	19.5 \\
    \ \ \ KDE ($S_I$)  &\textbf{24.1}	&\textbf{40.7}	&\textbf{24.3}	&\textbf{8.5}	&24.5	&48.3	&\textbf{23.0}	&\textbf{17.7}	&\textbf{15.1}	&\textbf{10.0}	&49.5	&17.5	&\textbf{41.0}	&19.4 \\
    \ \ \ KDE ($S_C$) & 24.0	&40.5&	\textbf{24.3}&	7.4&	\textbf{26.0}&	48.2	&22.5	&14.9	&14.0	&9.5	&49.7	&\textbf{20.6}	&\textbf{41.0}	&\textbf{19.9}\\
    \ \ \ Upp. Bound ($0.75$x)  & 27.6	&45.1	&28.2	&7.9	&30.8	&51.9	&29.7	&14.3	&21.5	&6.6	&54.4	&25.6	&44.7	&23.7 \\
    \ \ \ Upp. Bound (1x) &32.7	&51.9	&34.3	&14.4	&35.6	&51.8	&33.7	&21.1	&33.1	&5.7	&57.2	&36.7	&49.5	&24.6 \\
    \midrule
    \multicolumn{10}{@{}l}{\textbf{Without an Explicit Backward Label Mapping (Sec \ref{sec:no-back-map})}} \\
    \ \ \ KDE ($S_D$)&	5.4	&14.2&	3.7&	0.0&	0.9	&20.7	&3.2	&0.4	&1.2	&0.8	&27.9	&0.0	&5.3	&4.2 \\
    \ \ \ KDE ($S_I$)&6.1&	15.6&	4.0&	0.2	&0.8	&20.3	&2.3&	0.6&	0.7	&1.8	&30.8&	0.0	&7.0&	5.4 \\
    \ \ \ KDE ($S_C$)&6.0	&15.9&	3.8	&0.1&	0.9	&21.9	&3.0&	0.6	&0.9&	1.5&	30.2&	0.0	&6.7	&5.2 \\
    \midrule
    \multicolumn{10}{@{}l}{\textbf{Upper Bound with Ground Truth Saliency (Sec \ref{sec:perfect-saliency})}} \\
    \ \ \ KDE ($S_I$) &25.4	&42.6	&25.6&	9.1	&26.2	&49.5	&25.3	&17.4	&16.8	&10.1	&49.4	&23.4	&41.7	&19.4 \\
    \ \ \ KDE ($S_C$) &24.5&	41.7&	24.6&	7.5	&26.8&	48.8&	23.6&	14.5&	15.2&	9.7&	49.7	&22.6	&41.3&	19.8 \\
    \midrule
    \multicolumn{10}{@{}l}{\textbf{Sensitivity to Inter-Frame Motion (Sec \ref{sec:motion-sensitivity})}} \\
    \ \ \ KDE ($S_I$), $j=10$ &25.3&	42.9&	25.3	&8.4	&26.7&	49.1&	25.0	&16.4&	16.2	&10.1	&48.8	&25.0&	41.8&	19.5 \\
    \ \ \ KDE ($S_I$), $j=25$ &24.1	&41.0&	24.5	&6.4	&26.1	&49.0&	24.0&	12.6	&15.2	&9.0&	48.5&	22.9&	41.1&	19.6\\
    \ \ \ KDE ($S_I$), $j=50$ &22.5	&38.3&	22.9	&4.2&	24.1&	49.1&	21.9&	9.9	&14.4&	8.2	&48.4	&18.5&	39.0	&19.7\\
    \ \ \ KDE ($S_I$), $j=100$ &20.9&	35.1	&21.6&	2.8	&21.9&	48.0&	20.1&	7.1&	14.0	&6.8&	47.8&	15.3&	36.7&	19.1\\
    \ \ \ KDE ($S_I$), $j=200$ &20.0	&33.5&	20.6	&2.5&	20.5&	46.7&	19.2&	6.0	&13.4&	6.2	&46.7&	14.3&	35.5&	18.5\\
    \bottomrule
    \end{tabular}
    \end{adjustbox}
    Argoverse-HD after finetuning\\
    \begin{adjustbox}{width=1.0\linewidth,center}
    \begin{tabular}{@{}lcccccccccccccc@{}}
    \toprule
    Method    & AP & AP$_{50}$ & AP$_{75}$ & AP$_{S}$ & AP$_{M}$ & AP$_{L}$ & person & mbike & tffclight & bike & bus & stop & car & truck \\
    \midrule
    \multicolumn{10}{@{}l}{\textbf{Main Results} (copied from the main text for comparison)} \\
    \ \ \ Baseline &24.2	&38.9	&26.1	&4.9	&29.0	&50.9	&22.8	&7.5	&23.3	&5.9	&44.6	&19.3	&43.7	&26.6 \\
    \ \ \ Learned Sep. & 27.2&44.8&28.3&\textbf{12.2}&29.1&46.6&24.2&	14.0	&22.6&	7.7	&39.5&	\textbf{31.8}&	50.0&	27.8\\
    \ \ \ Learned Nonsep. & 25.9&42.9&26.5&10.0&28.4&48.5&25.2	&11.9	&20.9&	7.1&	39.5&	25.1&	49.4&	28.1\\
    \ \ \ KDE ($S_D$) & 26.7	&43.3&	27.8&	8.2	&29.7&	54.1	&25.4&	13.5&	22.0	&8.0	&\textbf{45.9}&	21.3&	48.1	&29.3\\
    \ \ \ KDE ($S_I$) & 28.0	& 45.5	&\textbf{29.2}	&10.4	&\textbf{31.0}	&\textbf{54.5}	&27.3	&16.9	&\textbf{24.3}	&\textbf{9.0}	&44.5	&23.2	&\textbf{50.5}	&28.4 \\
    \ \ \ KDE ($S_C$) &27.2&	44.7&	28.4	&9.1	&30.9&	53.6	&27.4	&14.5&	23.0&	7.0	&44.8	&21.9&	49.9&	\textbf{29.5} \\
    \ \ \ LKDE ($S_I$) & \textbf{28.1}	&\textbf{45.9}	&28.9	&10.3&	30.9	&54.1	&\textbf{27.5}&	\textbf{17.9}&	23.6&	8.1	&45.4	&23.1&	50.2&	28.7 \\
    \ \ \ Upp. Bound ($0.75$x) & 29.2	&47.6	&31.1	&11.6	&32.1	&53.3	&29.6	&12.7	&30.8	&7.9	&44.1	&29.8	&48.8	&30.1\\
    \ \ \ Upp. Bound (1x) & 33.3 & 53.9 & 35.0 & 16.8 & 34.8 & 53.6 & 33.1 & 20.9 & 38.7 & 6.7 & 44.7 & 36.7 & 52.7 & 32.7 \\
    \midrule
    \multicolumn{10}{@{}l}{\textbf{Without an Explicit Backward Label Mapping (Sec \ref{sec:no-back-map})}} \\
    \ \ \ KDE ($S_D$), no RoI GT &2.1	&2.6&	2.5	&0.0&	0.0&	4.0	&0.6&	0.0&	0.0&	0.6&	14.8&	0.0	&0.0&	0.9\\
    \ \ \ KDE ($S_D$) &1.8	&2.7	&1.9&	0.0&	0.0	&3.2&	0.6	&0.0&	0.0&	0.0	&13.3	&0.0	&0.1	&0.6\\
    \ \ \ KDE ($S_I$), no RoI GT &2.5&	3.0	&2.9	&0.0&	0.1	&4.3&	0.7	&0.0&	0.0&	0.6	&17.0	&0.9&	0.0	&0.9\\
    \ \ \ KDE ($S_I$) &2.0&	2.8&	2.4	&0.0&	0.0	&3.7&	0.6&	0.0	&0.0&	0.0	&14.8	&0.0	&0.3&	0.5\\
    \midrule
    \multicolumn{10}{@{}l}{\textbf{Upper Bound with Ground Truth Saliency (Sec \ref{sec:perfect-saliency})}} \\
    \ \ \ KDE ($S_I$) &29.6	&48.7	&30.7&	12.0	&32.8	&54.4	&28.3&	16.3&	27.7&	9.9	&43.9	&30.6	&50.9&	28.8 \\
    \ \ \ KDE ($S_C$) &27.8	&45.5&	28.8&	9.6	&31.7	&53.4	&27.5&	13.9&	24.7&	6.5	&44.5&	25.1	&50.2	&29.6 \\
    \midrule
    \multicolumn{10}{@{}l}{\textbf{Sensitivity to Inter-Frame Motion (Sec \ref{sec:motion-sensitivity})}} \\
    \ \ \ KDE ($S_I$), $j=10$ &29.4&	48.3&	30.7	&11.5	&32.8&	54.6&	27.9&	15.9&	27.2	&9.7&	43.7&	31.1&	50.6&	28.7 \\
    \ \ \ KDE ($S_I$), $j=25$ &28.0&	46.1&	29.2	&9.2&	32.1&	55.3&	26.4&	13.9&	25.9	&9.3&	43.9&	26.8&	49.2&	28.7\\
    \ \ \ KDE ($S_I$), $j=50$ &26.2	&42.9&	27.7	&6.6&	30.5&	54.9&	24.1&	12.1&	24.9	&8.6&	44.1&	21.8&	46.2&	27.9\\
    \ \ \ KDE ($S_I$), $j=100$ &24.5	&39.9&	25.8	&4.8	&28.6&	53.5&	22.3&	10.2&	23.5	&7.6	&43.5&	17.7	&43.9&	27.1\\
    \ \ \ KDE ($S_I$), $j=200$ &23.6	&38.3&	25.2	&4.2&	27.8&	53.0	&21.4&	8.6	&22.8	&7.4	&42.9&	16.6&	42.7&	26.6\\
    \bottomrule
    \end{tabular}
    \end{adjustbox}
    \vspace{-0.5em}
    \caption{Additional diagnostics experiments on Argoverse-HD. Please refer to Sec~\ref{app:diagnostics} for a detailed discussion.}
    \label{tab:additional-diagnostics}
\end{table*}

Having noted that the $S_I$ and $S_C$ formulations are sensitive to the accuracy of the previous-frame detections, in this section, we further test its robustness to motion between frames. We use ground truth bounding boxes (rather than detections) from the previous frame in order to isolate the effect of motion on accuracy. We introduce a jitter parameter $j$ and
translate each of the ground truth bounding boxes in the x and y directions by values sampled from $\mathcal{U}(-j,j)$. The translation values are in pixels in reference to the original image size of $1920\times 1200$. As in Sec~\ref{sec:perfect-saliency}, this is a purely testing-time change. Also note that the upper bound experiments in Sec~\ref{sec:perfect-saliency} follows by setting $j=0$. We test only on $S_I$ and report the full results in Tab \ref{tab:additional-diagnostics}. We also plot summarized results and discuss observations in \ref{fig:motion-sensitivity}.

\begin{figure}[t]
\centering
\includegraphics[width=\linewidth]{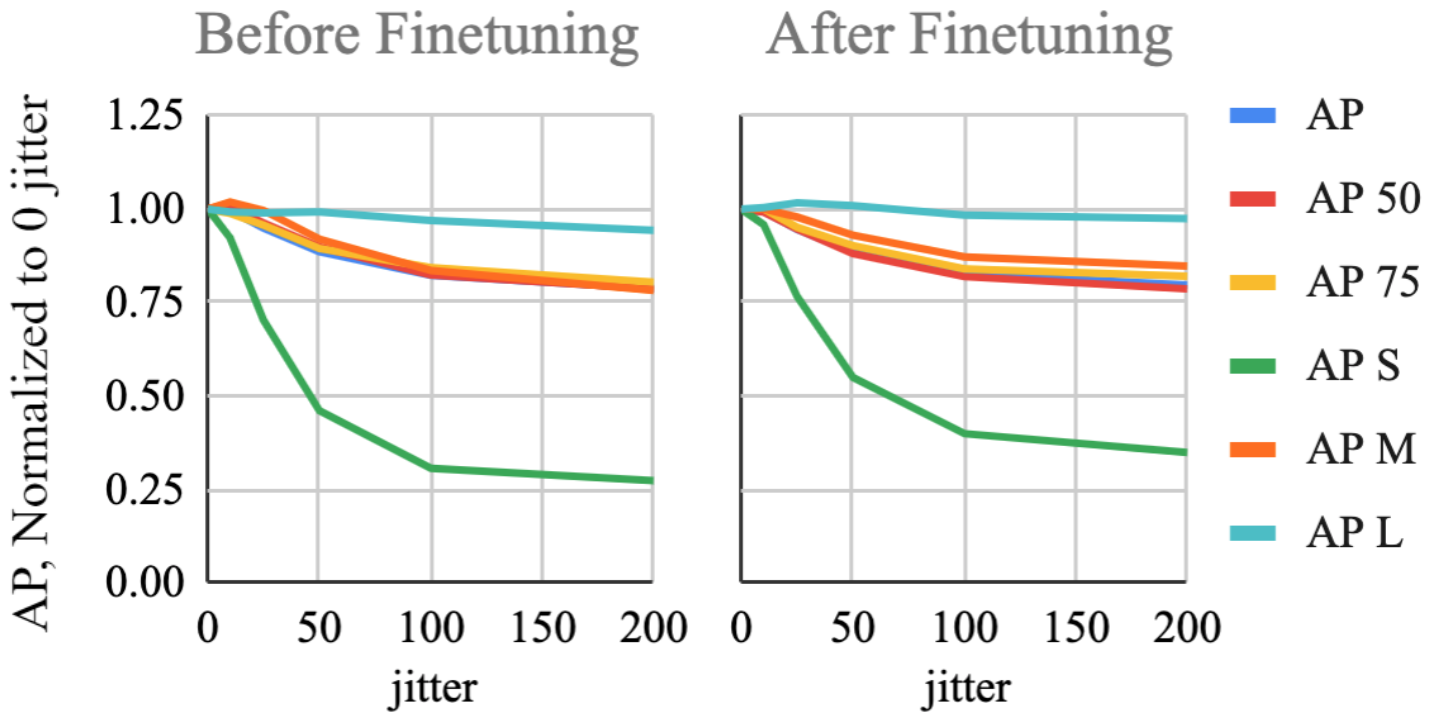}
\vspace{-1.5em}
\caption{
Plots showing the effect of motion (jitter) on AP using the KDE $S_I$ formulation. Results have been normalized according to the AP at 0 jitter. As is intuitive, motion affects AP$_S$ the most and AP$_L$ the least. After finetuning (with an artificial jitter of 50), we see that the model reacts less adversely to jitter, indicating that our regularization has helped.}
\label{fig:motion-sensitivity}
\end{figure}

\section{FOVEA Beyond Faster R-CNN}
\label{app:other-det}

In the main text and other sections of the appendix, we conduct our experiment based on Faster R-CNN. However, our proposed warping-for-detection framework is agnostic to specific detectors. To show this, we test our methods on RetinaNet \cite{lin2017focal}, a popular single-stage object detector, and on YOLOF \cite{chen2021you}, a recent YOLO variant that avoids bells and whistles and long training schedules (up to 8x for ImageNet and 11x for COCO compared to standard schedules for YOLOv4~\cite{bochkovskiy2020yolov4}).

For both these detectors, we test baselines at $0.5$x and $0.75$x scales both before and after finetuning. We then compare these results against our KDE $S_I$ method at $0.5$x scale. 
We use a learning rate of 0.01 for the RetinaNet KDE $S_I$ model and 0.005 for the RetinaNet baselines. All other training settings for RetinaNet are identical to the Faster-RCNN baseline. For YOLOF, we use a learning rate of 0.012 and keep all other settings true to the original paper. Results are presented in Tab~\ref{tab:another-det}.

\begin{table}[t]
    \centering
    \begin{adjustbox}{width=1.0\linewidth,center}
    \begin{tabular}{@{}lcccccc@{}}
    \toprule
    Method    & AP & AP$_{50}$ & AP$_{75}$ & AP$_{S}$ & AP$_{M}$ & AP$_{L}$ \\
    \midrule
    \multicolumn{6}{@{}l}{\textbf{RetinaNet, Before Finetuning on Argoverse-HD}}\\
    Baseline ($0.5$x) &18.5&	29.7&	18.6&	1.3	&17.2	&48.8 \\
    KDE ($S_I$) & 18.5&	31.2&	17.9&	4.5	&16.8	&44.9\\
    Upp. Bound ($0.75$x) & 24.8&	38.8&	25.5&	4.5	&28.7	&52.0\\
    \midrule
    \multicolumn{6}{@{}l}{\textbf{RetinaNet, After Finetuning on Argoverse-HD}}\\
    Baseline ($0.5$x) & 22.6	&38.9&	21.4&	4.0	&22.0	&53.1\\
    KDE ($S_I$) & 24.9	&40.3	&25.3&	7.1&	27.7	&50.6\\
    Upp. Bound ($0.75$x) & 29.9	&48.6&	30.1&	9.7	&32.5	&54.2\\
    \midrule
    \multicolumn{6}{@{}l}{\textbf{YOLOF, Before Finetuning on Argoverse-HD}}\\
    Baseline ($0.5$x) & 15.0 & 25.4 & 14.3 & 0.6 & 11.0 & 46.0 \\
    KDE ($S_I$) & 16.8 & 29.0 & 16.0 & 0.9 & 14.0 & 46.4 \\
    Upp. Bound ($0.75$x) & 21.6	&35.5&	22.3&	2.3&	22.2&	52.7\\
    \midrule
    \multicolumn{6}{@{}l}{\textbf{YOLOF, After Finetuning on Argoverse-HD}}\\
    Baseline ($0.5$x) & 18.4&	30.5&	18.3&	1.4&	16.5&	47.9\\
    KDE ($S_I$) & 21.3&	36.7&	20.2&	3.5&	21.8&	49.7\\
    Upp. Bound ($0.75$x) & 25.1&	41.3&	25.3&	4.7&	27.6&	54.1\\
    \bottomrule
    \end{tabular}
    \end{adjustbox}
    \caption{Experiments with RetinaNet \cite{lin2017focal} and YOLOF \cite{chen2021you}. We follow the same setup as the experiment with Faster R-CNN. The top quarter suggests that unlike Faster R-CNN, RetinaNet does not work off-the-shelf with our KDE warping. However, the second quarter suggests similar performance boosts as with Faster R-CNN can be gained after finetuning on Argoverse-HD. Interestingly, for YOLOF, our method boosts AP in all categories -- small, medium, and large -- even with off-the-shelf weights.}
    \label{tab:another-det}
\end{table}

\section{Comparison Against Additional Baselines}
\label{app:other-baselines}

There are other approaches that make use of image warping or patch-wise zoom for visual understanding. The first noticeable work \cite{recasens2018learning}, explained extensively in the main text, warps the input image for tasks that have labels invariant to warping. The second noticeable work \cite{gao2018dynamic} employs reinforcement learning (RL) to decide which patches to zoom in for high-resolution processing. In this section, we attempt to compare our FOVEA with these two approaches.

Our method builds upon spatial transformer networks~\cite{jaderberg2015spatial,recasens2018learning} and we have already compared against \cite{recasens2018learning} sporadically in the main text. Here provides a summary of all the differences (see Tab~\ref{tab:techcontrib}). A naive approach might
directly penalize the discrepancy between the output of the (warped) network and the unwarped ground-truth
in an attempt to implicitly learn the inverse mapping,
but this results in abysmal performance
(dropping 28.1 to 2.5 AP, discussed in Sec~\ref{sec:no-back-map}). To solve this issue, in Sec~\ref{sec:background}, we note that \cite{jaderberg2015spatial,recasens2018learning} actually learn a backward map $\T^{-1}$ instead of a forward one $\T$. This allows us to add a backward-map layer that transforms bounding box coordinates back to the original space via $\T^{-1}$, dramatically improving accuracy. 
A second significant difference with \cite{jaderberg2015spatial,recasens2018learning} is our focus on attention-for-efficiency. If the effort required to determine where to attend is more than the effort to run the raw detector, attentional processing can be inefficient (see the next paragraph). \cite{recasens2018learning} introduces a lightweight saliency network to produce a heatmap for where to attend; however, this model does not extend to object detection, perhaps because it requires the larger capacity of a detection network (see Sec~\ref{Baseline and Setup}). Instead, we replace this feedforward network with an essentially {\em zero}-cost saliency map constructed via a simple but effective global spatial prior (computed offline) or temporal prior (computed from previous frame's detections).
Next, we propose a technique to prevent cropping during warping (via reflection padding, as shown in Fig~\ref{fig:anticropping}), which also boosts performance by a noticeable amount. Finally, as stated in the training formulation in Sec~\ref{sec:warping4det}, it {\em doesn't even make sense} to train a standard RPN-based detector with warped input due to choice of delta encoding (which normally helps stabilize training). We must remove this standard encoding and use GIoU to compensate for the lost stability during training.

\begin{table}[!h]
\small
\centering
\adjustbox{width=1\linewidth}{
\begin{tabular}{lcc}
\toprule
Method                                    & AP            \\ \midrule
FOVEA (Ours full)                               & 28.1          \\
w/o Explicit backward mapping             & 2.5           \\
w/o KDE saliency (using saliency net as in \cite{recasens2018learning}) & Doesn't train \\
w/o Anti-crop regularization              & 26.9             \\
w/o direct RPN box encoding   & N/A             \\
\bottomrule
\end{tabular}
}
\caption{Summary of key modifications in FOVEA.}
\label{tab:techcontrib}
\end{table}

Next, we attempt to compare against this RL-based zoom method \cite{gao2018dynamic} using our baseline detector (public implementation from mmdetection~\cite{mmdetection}) on their Caltech Pedestrian Dataset~\cite{Dollar2012PAMI}. However, while their full-scale $800 \times 600$ Faster R-CNN detector reportedly takes 304ms, our implementation is {\em dramatically} faster (44ms), consistent with the literature for modern implementations and GPUs.
This changes the conclusions of that work because full-scale processing is now faster than coarse plus zoomed-in processing (taking 28ms and 25ms respectively), even assuming a zero-runtime RL module (44ms $<$ 28ms + 25ms).

\section{Additional Visualizations}

Please refer to Fig \ref{fig:kde-supp-visuals} and \ref{fig:kde-visuals} for additional qualitative results of our method.

\begin{figure*}[t]
\centering
\includegraphics[width=\linewidth]{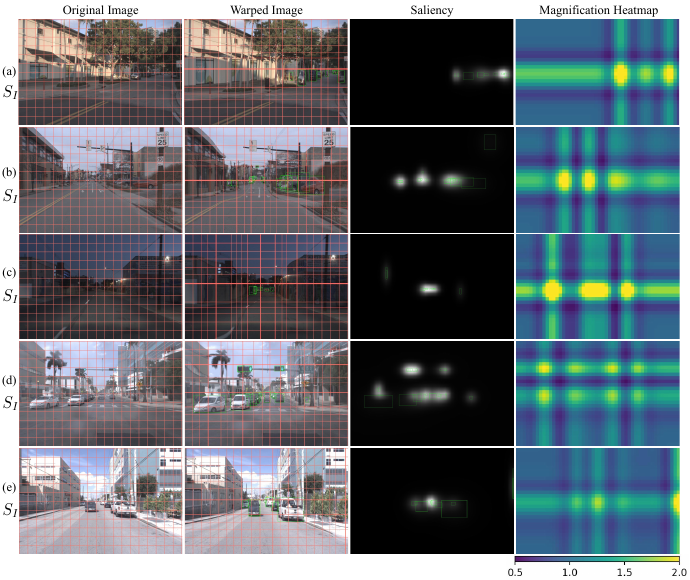}
\vspace{-1.5em}
\caption{Additional examples of the $S_I$ KDE warping method. Bounding boxes on the saliency map denote previous frame detections, and bounding boxes on the warped image denote current frame detections. The magnification heatmap depicts the amount of magnification at different regions of the warped image. 
(a) is an example of $S_I$ correctly adapting to an off-center horizon. 
(b) shows a multimodal saliency distribution, leading to a multimodal magnification in the $x$ direction.
(c) is another example of $S_I$ correctly magnifying small objects in the horizon.
(d) is a failure case in which duplicate detections of the traffic lights in the previous frame leads to more magnification than desired along that horizontal strip. One solution to this could be to weight our KDE kernels by the confidence of the detection.
(e) is another failure case of $S_I$, in which a small clipped detection along the right edge leads to extreme magnification in that region. 
One general issue we observe is that the regions immediately adjacent to magnified regions are often contracted. This is visible in the magnification heatmaps as the blue shadows around magnified regions. This is a byproduct of the dropoff in attraction effect of the local attraction kernel. Perhaps using non-Gaussian kernels can mitigate this issue.
}
\label{fig:kde-supp-visuals}
\end{figure*}

\begin{figure*}[t]
\centering
\includegraphics[width=\linewidth]{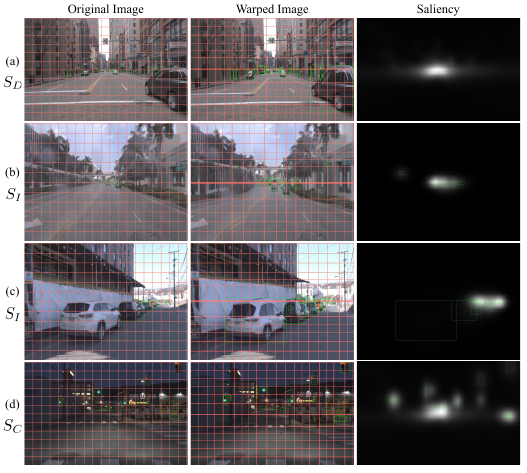}
\vspace{-1em}
\caption{Examples of KDE warp computed from bounding boxes, extracted from a training dataset ($S_D$) or the previous frame's detections ($S_I, S_C$). We visualize predicted bounding boxes in the warped image. Recall that large objects won't be visible in the saliency due to their large variance from Eq~\ref{eq:kde}. (a) $S_D$ magnifies the horizon (b) $S_I$ magnifies the center of the image, similar to $S_D$ (c) $S_I$ adapts to magnify the mid-right region (d) $S_C$'s saliency combines the temporal and spatial biases.}
\label{fig:kde-visuals}
\end{figure*}

\section{Detection-Only Streaming Evaluation}
\label{app:det-streaming}

In 
\ifstandalonesupplement
    Sec~4.2
\else
    Sec~\ref{sec:streaming}
\fi
of the main text, we provide the full-stack evaluation for streaming detection. Here we provide the detection-only evaluation for completeness in Tab~\ref{tab:streaming-det}. This setting only allows detection and scheduling, and thus isolating the contribution of tracking and forecasting. We observe similar trend as in the full-stack setting in 
\ifstandalonesupplement
    Tab~2.
\else
    Tab~\ref{tab:streaming-full}.
\fi

\section{Additional Implementation Details}
\label{app:impl-details}

In this section, we provide additional details necessary to reproduce the results in the main text. 

For the learned separable model from Sec~\ifstandalonesupplement 4.1.2\else\ref{direct saliency experiments}\fi, we use two arrays of length $31$ to model saliency along the $x$ and $y$ dimensions, and during training, we blur the image with a $47\times47$ Gaussian filter in the first epoch, a trick introduced in \cite{recasens2018learning} to force the model to zoom. 
For the learned nonseparable model, we use an $11\times11$ saliency grid, and we blur the image with a $31\times31$ filter in the first epoch.
We use an attraction kernel $k$ with a standard deviation of $5.5$ for both versions. Additionally, we multiply the learning rate and weight decay of saliency parameters by 0.5 in the first epoch and 0.2 in the last two epochs, for stability. We find that we don't need anti-crop regularization here, because learning a fixed warp tends to behave nicely.

For each of our KDE methods, we use arrays of length $31$ and $51$ to model saliency in the vertical and horizontal directions, respectively. This is chosen to match the aspect ratio of the original input image and thereby preserve the vertical and horizontal ``forces" exerted by the attraction kernel.

For the baseline detector, we adopt the Faster R-CNN implementation of mmdetection 2.7 \cite{mmdetection}. All our experiments are conducted in an environment with PyTorch 1.6, CUDA 10.2 and cuDNN 7.6.5. For streaming evaluation, we mention a performance boost due to better implementation in Tab~\ref{tab:streaming-det}
\&
\ifstandalonesupplement
    Tab~2,
\else
    Tab~\ref{tab:streaming-full},
\fi
and the changes are mainly adopting newer versions of mmdetection and cuDNN compared to the solution in \cite{Li2020StreamingP} (switching from a smooth L1 loss to L1 loss for the regression part and code optimization).

\begin{table}[t]
\small
\centering
\begin{tabular}{clcccc}
\toprule
ID & Method    & AP            & AP$_S$       & AP$_M$                  & AP$_L$                   \\
\midrule
1  & Prior art \cite{Li2020StreamingP}       & 13.0          & 1.1          & 9.2 & 26.6 \\
\midrule
2  & + Better implementation                                 & 14.4          & 1.9          & 11.5                    & \textbf{27.9}                     \\
3  & + Train with pseudo GT                       & 15.7          & 3.0          & 14.8                    & 27.1                     \\
\midrule
4  & 2 + Ours ($S_I$)                                            & 15.7          & 4.7          & 12.8                    & 26.8                     \\
5  & 3 + Ours ($S_I$)                                            & \textbf{17.1} & \textbf{5.5} & \textbf{15.1}           & 27.6           \\
\bottomrule
\end{tabular}
\caption{Streaming evaluation in the detection-only setting. First, we are able to improve over previous state-of-the-art through better implementation (row 2) and training with pseudo ground truth (row 3). Second, our proposed KDE warping further boosts the streaming accuracy (row 4-5).}
\label{tab:streaming-det}
\end{table}

\else
    \maketitle
    \begin{abstract}
Efficient processing of high-res video streams is safety-critical for many robotics applications such as autonomous driving. To maintain real-time performance, many practical systems downsample the video stream. But this can hurt downstream tasks such as (small) object detection. Instead, we take inspiration from biological vision systems that allocate more foveal ``pixels" to salient parts of the scene. We introduce FOVEA, an approach for intelligent downsampling that ensures salient image regions remain ``magnified" in the downsampled output. Given a high-res image, FOVEA applies a differentiable resampling layer that outputs a small fixed-size image canvas, which is then processed with a differentiable vision module (e.g., object detection network), whose output is then differentiably backward mapped onto the original image size. The key idea is to resample such that background pixels can make room for salient pixels of interest. In order to ensure the overall pipeline remains efficient, FOVEA makes use of cheap and readily available cues for saliency, including dataset-specific spatial priors or temporal priors computed from object predictions in the recent past.  
On the autonomous driving datasets Argoverse-HD and BDD100K, our proposed method boosts the detection AP over standard Faster R-CNN, both with and without finetuning. Without any noticeable increase in compute, we improve accuracy on small objects by over 2x without degrading performance on large objects. Finally, 
FOVEA sets a new record for streaming AP (from 17.8 to 23.0 on a GTX 1080 Ti GPU), a metric designed to capture both accuracy and latency.


 
\end{abstract}
    \section{Introduction}

\begin{figure}[t]
\centering
\includegraphics[width=1\linewidth]{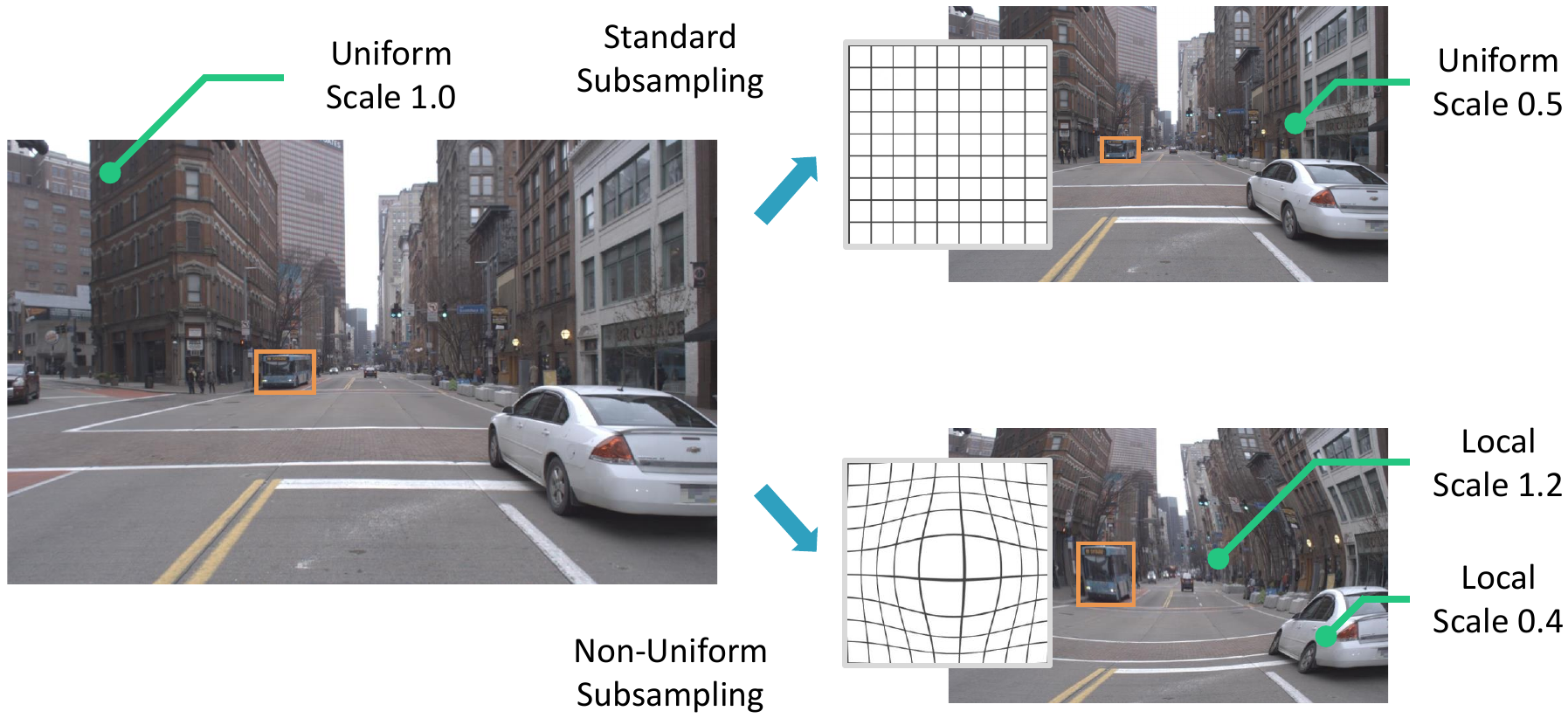}
\caption{Standard image downsampling (top right) limits the capability of the object detector to find small objects. In this paper, we propose an attentional warping method (bottom right) that enlarges salient objects in the image while maintaining a small input resolution. Challenges arise when warping also alters the output labels (\eg, bounding boxes).}
\label{fig:teaser-teaser}
\end{figure}

Safety-critical robotic agents such as self-driving cars make use of an enormous suite of high-resolution perceptual sensors, with the goal of minimizing blind spots, maximizing perception range, and ensuring redundancy~\cite{Argoverse,nuscenes2019,Waymo}. 
We argue that ``over-sensed'' perception platforms provide unique challenges for vision algorithms since those visual sensors must rapidly consume sensor streams while continuously reporting back the state of the world.
While numerous techniques exist to make a particular model run fast, such as quantization~\cite{Vanhoucke2011ImprovingTS}, model compression~\cite{Cheng2017ASO}, and inference optimization~\cite{RaganKelley2013HalideAL}, at the end of the day, simple approaches that subsample sensor data (both spatially by frame downsampling and temporally by frame dropping) are still most effective for meeting latency constraints~\cite{Li2020StreamingP}.
However, subsampling clearly throws away information, negating the goals of high-resolution sensing in the first place! 
This status quo calls for novel vision algorithms.

%
To address this challenge, we take inspiration from the human visual system; biological vision makes fundamental use of {\em attentional} processing.
While current sensing stacks make use of regular grid sampling, 
the human vision system in the periphery has a much lower resolution than in the center (fovea), due to the pooling of information from retinal receptors by retinal ganglion cells. Such variable resolution is commonly known as foveal vision \cite{Larson2009TheCO}.

In this paper, we propose FOVEAted image magnification (FOVEA) for object detection, which retains high resolution for objects of interest while maintaining a small canvas size. We exploit the sparsity of detection datasets -- objects of interest usually only cover a portion of the image. {\em The key idea is to resample such that background pixels can make room for salient pixels of interest.
} The input images are downsampled and warped such that
salient
areas in the warped image have higher resolutions. While image warping has been explored for image classification \cite{jaderberg2015spatial,recasens2018learning} and regression \cite{recasens2018learning}, major challenges remain when applying such methods to detailed spatial prediction tasks such as object detection. First, processing warped images will produce warped spatial predictions (bounding box coordinates). We make use of 
differentiable backward maps 
to unwarp spatial predictions back to the original space. Second, it is hard to efficiently identify salient regions; in the worst case, a saliency network tuned for object detection may be as expensive as the downstream detection network itself, thereby eliminating any win from downsampling. 
%
In our case, we make use of cheap and readily available saliency cues, either in the form of dataset-specific spatial priors (i.e., small objects tend to exist near a fixed horizon) or temporal priors (small objects tend to lie nearby small object predictions from previous frames).
Third, previous image warps (tuned for image classification tasks) can produce cropped image outputs. Since objects can appear near the image boundary, we introduce anti-cropping constraints on the warping. 

We validate our approach on two self-driving datasets for 2D object detection: Argoverse-HD \cite{Li2020StreamingP} and BDD100K \cite{bdd100k}. First, we show that FOVEA 
 can improve the performance of off-the-shelf detectors (Faster R-CNN \cite{ren2015faster}). Next, we finetune detectors with differentiable image warping and backward label mapping, further boosting performance. In both cases, small objects improve by more than 2x in average precision (AP). Finally, we evaluate FOVEA under streaming perception metrics designed to capture both accuracy and latency~\cite{Li2020StreamingP}, producing state-of-the-art results. 

\section{Related Work}
\label{sec:related}

\paragraph{Object detection}
Object detection is one of the most fundamental problems in computer vision. Many methods have pushed the state-of-the-art in detection accuracy~\cite{girshick2014rich,ren2015faster,lin2017feature,chen2019hybrid,qiao2020detectors}, and many others aim for improving the efficiency of the detectors~\cite{Liu2016SSDSS,redmon2018yolov3,tan2020efficientdet,bochkovskiy2020yolov4}. The introduction of fully convolution processing~\cite{sermanet2013overfeat} and spatial pyramid pooling~\cite{He2015SpatialPP} have allowed us to process the input image in its original size and shape. However, it is still a common practice to downsample the input image for efficiency purposes. Efficiency becomes a more prominent issue when people move to the video domain. In video object detection, the focus has been on how to make use of temporal information to reduce the number of detectors invoked~\cite{zhu2017flow,zhu2018towards,luo2019detect}. These methods work well on simple datasets like ImageNet~VID~\cite{ILSVRC15}, but might be unsuitable for the self-driving car senarios, where multiple new objects appear at almost every frame. Furthermore, those methods are usually designed to work in the offline fashion, \ie, allowing access to future frames. Detection methods are the building blocks of our framework, and our proposed approach is largely agnostic to any particular detector. 

\paragraph{Online/streaming perception} 

In the online setting, the algorithm must work without future knowledge. \cite{lin2019tsm} proposes the Temporal Shift Module that enables video understanding through channel shifting and in the online setting, the shifting is restricted to be uni-directional. \cite{Bergmann2019TrackingWB} proposes a multi-object tracking method that takes input previous frame detection as addition proposals for the current frame. Our method also takes previous frame detection as input, but we use that to guide image warping. Streaming accuracy~\cite{Li2020StreamingP} is a recently proposed metric that evaluates the output of a perception algorithm at all time instants, forcing the algorithm to consider the amount of streaming data that must be ignored while computation is occuring. 
\cite{Li2020StreamingP} demonstrates that streaming object detection accuracy can be significantly improved by tuning the input frame resolution and framerate. In this work, we demonstrate that adaptive attentional processing is an orthogonal dimension for improving streaming performance. 

\begin{figure*}[t]
\centering
\includegraphics[width=0.8\linewidth]{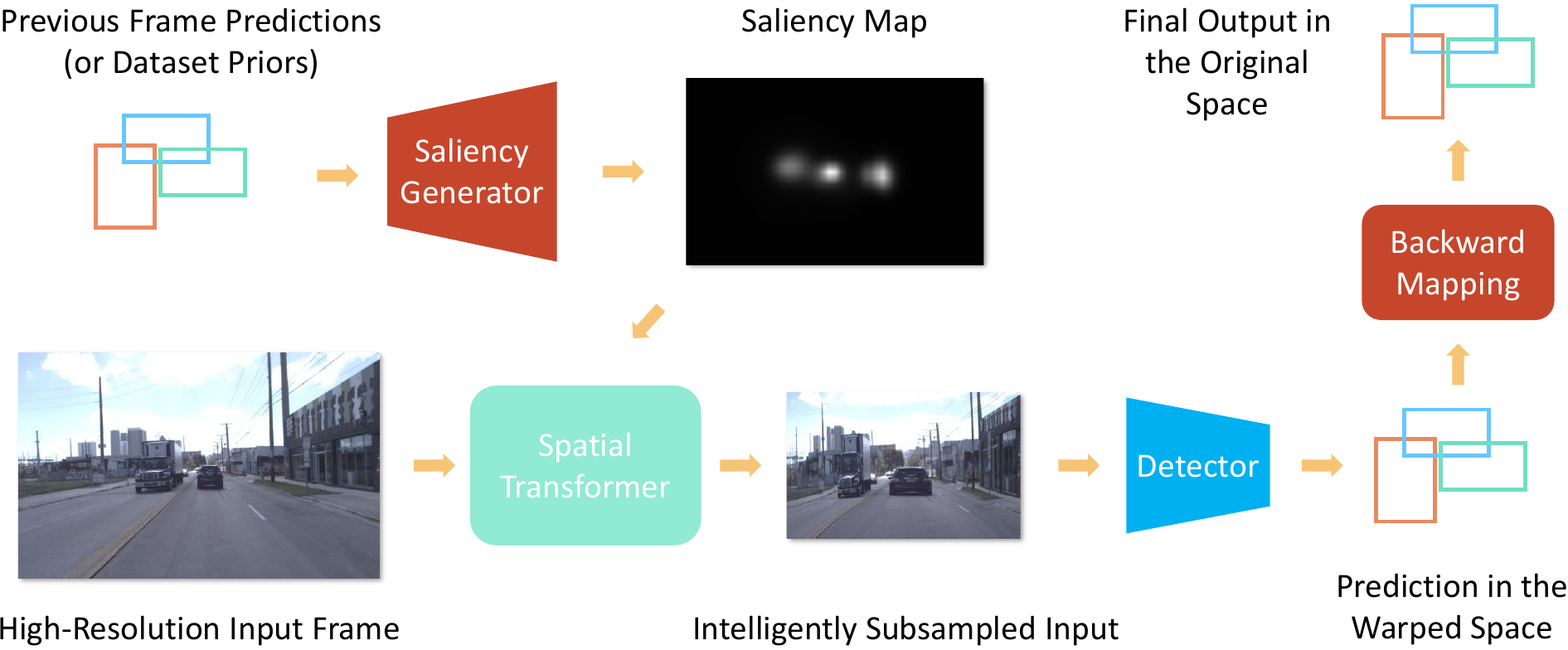}
\caption{Our proposed method for object detection. Given bounding box predictions from the previous frame (if the input are videos) or a collection of all the ground truth bounding boxes in the training set, the saliency generator creates a saliency map and that is fed into the spatial transformer (adapted from \cite{recasens2018learning,jaderberg2015spatial}) to downsample the high-resolution input frame while magnifying salient regions. Then we feed the downsampled input into a regular object detector, and it produces bounding box output in the warped space, which is then converted back to the original image space as the final output.
}
\label{fig:teaser}
\end{figure*}

\paragraph{Adaptive visual attention} Attentional processing has been well studied in the vision community, and it appears in different forms \cite{dai2017deformable,huang2017learning,kirillov2020pointrend,liu2018dynamic,li2017not,verelst2020dynamic}. Specially in this paper, we focus on dynamic resolutions. For image classification, \cite{uzkent2020learning} designs an algorithm to select high-resolution patches, assuming each patch is associated with a data acquisition cost. \cite{marin2019efficient} applies non-uniform downsampling to semantic segmentation and relies on the network to learn both the forward and backward mapping, whose consistency is not guaranteed. For object detection, a dynamic zoom-in algorithm is proposed that processes high-resolution patches sequentially \cite{gao2018dynamic}. However, sequential execution might not meet latency requirements for real-time applications. Most similar to our work, \cite{recasens2018learning} proposes an adaptive image sampling strategy that allocates more pixels for salient areas, allowing a better downstream task performance. But the method only works for image classification and regression, where the output is agnostic to the input transformation.

\section{Approach}

Assume we are given a training set of image-label pairs $(I,L)$. We wish to learn a nonlinear deep predictor $f$ that produces a low loss $\mathcal L(f(I),L)$. Inspired by past work~\cite{recasens2018learning,jaderberg2015spatial}, we observe that certain labeling tasks can be performed more effectively by warping/resampling the input image. However, when the label $L$ itself is spatially defined (e.g., bounding box coordinates or semantic pixel labels), the label itself may need to be warped, or alternatively, the output of the deep predictor may need to be inverse-warped.

In this section, we first introduce the saliency-guided spatial transform from related work as the foundation of our method. Next, we introduce our solutions to address the challenges in image warping for object detection. An overview of FOVEA, our method, is shown in Fig~\ref{fig:teaser}.

\subsection{Background: Saliency-Guided Spatial Transform}
\label{sec:background}

\begin{figure}[t]
\centering
\includegraphics[width=0.9\linewidth]{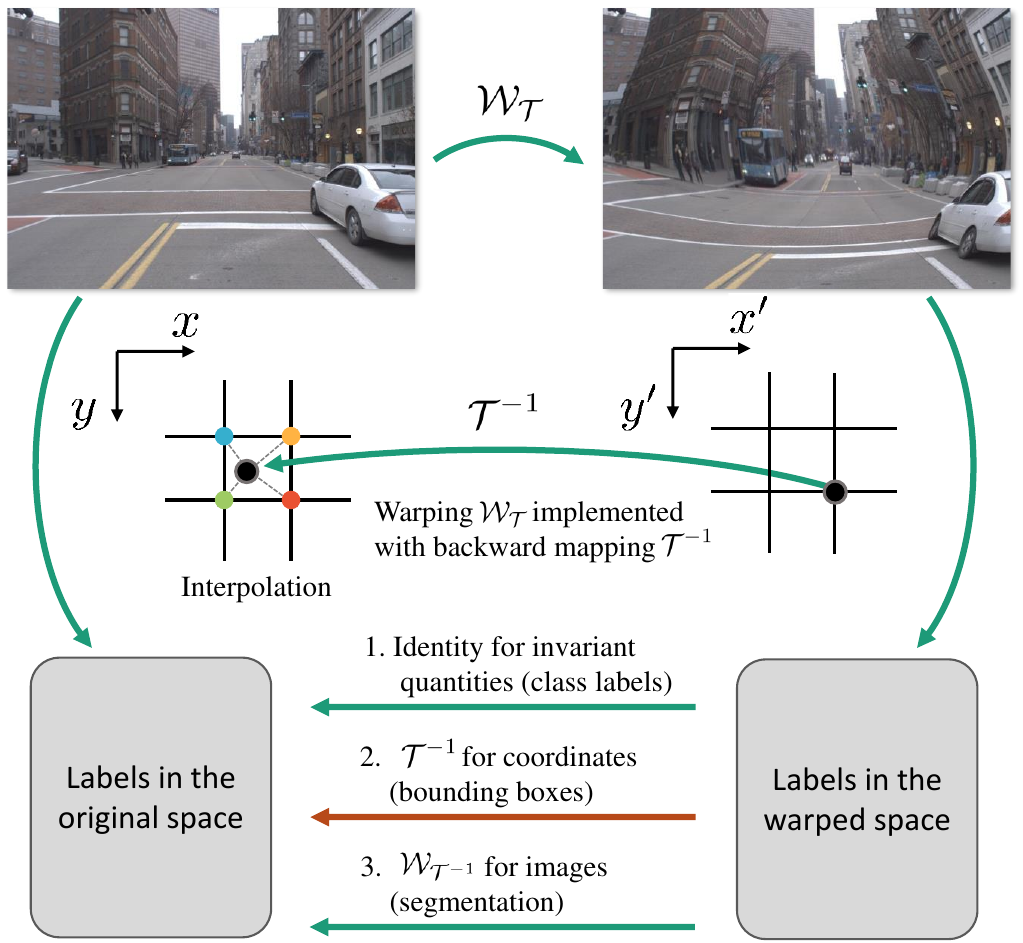}
\caption{Image warps $\W_{\T}$ are commonly implemented via a backward map $\T^{-1}$ followed by (bilinear) interpolation of nearby source pixel grid values, since forward mapping $\T$ can result in target pixel positions that do not lie on the pixel grid (not shown). Though image warping is an extensively studied topic (notably by ~\cite{jaderberg2015spatial,recasens2018learning} in the context of differentiable neural warps), its effect on labels is less explored because much prior art focuses on global labels invariant to warps (e.g. an image class label). We explore warping for spatial prediction tasks whose output must be transformed back into the original image space to generate consistent output. Interestingly, transforming pixel-level labels with warp $\W_{\T^{-1}}$ requires inverting $\T^{-1}$, which can be difficult depending on its parameterization~\cite{beier1992feature}. In this paper, we focus on transforming pixel {\em coordinates} of bounding boxes, which requires only the already-computed backward map $\T^{-1}$ (the red arrow).
}
\label{fig:warping}
\end{figure}

The seminal work of spatial transformer networks (STN) introduces a differentiable warping layer for input images and feature maps \cite{jaderberg2015spatial}. It was later extended to incorporate a saliency map to guide the warping \cite{recasens2018learning}. Here we provide implementation details that are crucial to our method. Please refer to the original papers \cite{jaderberg2015spatial,recasens2018learning} for more details.

A 2D transformation can be written as:
\begin{align}
    \T: (x, y) \to (x', y'),
\end{align}
where $(x, y)$ and $(x', y')$ are the input and output coordinates. Since image pixels are usually discrete, interpolation is required to sample values at non-integral coordinates. An image warp $\W_{\T}$ takes input an image $I$, samples the pixel values according to the given transformation $\T$, and outputs the warped image $I'$:
\begin{align}
I'(\T(x, y)) = I(x,y)
\end{align}
Naive forward warping of discrete pixel locations from input $I$ can result in non-integral target pixel positions that need to be ``splatted" onto the pixel grid of $I$, which can produce artifacts such as holes. Instead, image warps are routinely implemented via a {\em backward map}~\cite{beier1992feature}: iterate over each output pixel grid location, compute its {\em inverse mapping} $\T^{-1}$ to find its corresponding input coordinates (which may be non-integral), and bilinearly interpolate its color from neighboring input pixel grid points:
\begin{align}
I'(x, y) = I(\T^{-1}(x, y))
\end{align}
{\em In other words, the implementation of $\W_{\T}$ only requires the knowledge of the inverse transformation $\T^{-1}$}. The pixel iteration can be replaced with a batch operation by using a grid generator and apply the transformation $\T^{-1}$ over the entire grid.

STN uses a differentiable formulation of $\T^{-1}_{\theta}$ (parameterized by $\theta$) and an ensuing bilinear grid sampler, which is differentiable and parameter-free. \cite{recasens2018learning} proposes a special form of $\T^{-1}$ parameterized by a saliency map $S$: $\T^{-1}_{\theta} = \T^{-1}_{S}$. This transform has a convolution form and is therefore fast, using the intuition that each input pixel $(x,y)$ attracts samples from the original image with a force $S(x,y)$, leading to more sampling at salient regions. {\em We point out that both \cite{jaderberg2015spatial} and \cite{recasens2018learning} ignore the effect of warping on the output label space and skip the modeling of the forward transform $\T$, which (we will show) is required for unwarping certain label types.}

\subsection{Image Warping for Object Detection}
\label{sec:warping4det}

In this section, we first explain our high-level inference formulation, then our specific form of the warping, and in the end some adjustments for training the task network.

\paragraph{Inference formulation}
We visually lay out the space of image and label warps in 
\ifunlimited
Fig~\ref{fig:warping}.
Recent methods for differentiable image warping assume labels are invariant under the warping (the first pathway in Fig~\ref{fig:warping}). For object detection, however, image warping clearly warps bounding box outputs. To produce consistent outputs (e.g., for computing bounding box losses during learning), these warped outputs need to transformed back into the original space (the second pathway in Fig~\ref{fig:warping}).
\else
Fig~\ref{fig:warping}, which shows that we have to transform the bounding box outputs accordingly after the warping. 
\fi
Quite conveniently, because standard image warping is implemented via the backward map $\T^{-1}$, the backward map is already computed in-network and so can be directly applied to the pixel coordinates of the predicted bounding box.
The complete procedure for our approach $\hat{f}$ can be written as $\hat{f}(I, \T)=\T^{-1}(f(\W_\T(I)))$. where $f(\cdot)$ is the nonlinear function that returns bounding box coordinates of predicted detections. Importantly, this convenience doesn't exist when warping pixel-level {\em values}; \eg, when warping a segmentation mask back to the original image input space (the third pathway in Fig~\ref{fig:warping}). Here, one needs to invert $\T^{-1}$ to explicitly compute the forward warp $\T$.

\paragraph{Warping formulation}

We adopt the saliency-guided warping formulation from \cite{recasens2018learning}:
\begin{align}
    \T^{-1}_x(x,y) = \frac{\int_{x', y'} S(x', y') k((x,y), (x', y')) x'}{\int_{x', y'} S(x',y')k((x,y), (x',y'))}, \\
    \T^{-1}_y(x,y) = \frac{\int_{x', y'} S(x', y') k((x,y), (x', y')) y'}{\int_{x', y'} S(x',y')k((x,y), (x',y'))},
\end{align}
where $k$ is a distance kernel (we use a Gaussian kernel in our experiments). 
However, in this general form, axis-aligned bounding boxes might have different connotations in the original and warped space. To ensure axis-alignment is preserved during the mapping, we restrict the warping to be separable along the two dimensions, \ie, $\T^{-1}(x, y) = (\T^{-1}_x(x), \T^{-1}_y(y))$.
For each dimension, we adapt the previous formulation to 1D:
\begin{align}
\label{torralba warp}
    \T^{-1}_x(x) = \frac{\int_{x'} S_x(x')k(x', x) x'}{\int_{x'} S_x(x')k(x, x')}, \\
    \T^{-1}_y(y) = \frac{\int_{y'} S_y(y')k(y', y)y'}{\int_{y'} S_y(y')k(y, y')}.
\end{align}
We call this formulation \emph{separable} and the general form \emph{nonseparable}. Note that the nonseparable formulation has a 2D saliency map parameter, whereas the separable formulation has two 1D saliency maps, one for each axis.
Fig~\ref{fig:separable} shows an example of each type of warp.

One nice property of $\T^{-1}$ is that it is differentiable and thus can be trained with backpropagation. One limitation though is that its inverse $\T$ doesn't have a closed-form solution, nor does its derivative. The absence of $\T$ is not ideal, and we propose some workaround as shown in the following subsection. 

\begin{figure}[t]
\centering
\includegraphics[width=0.8\linewidth]{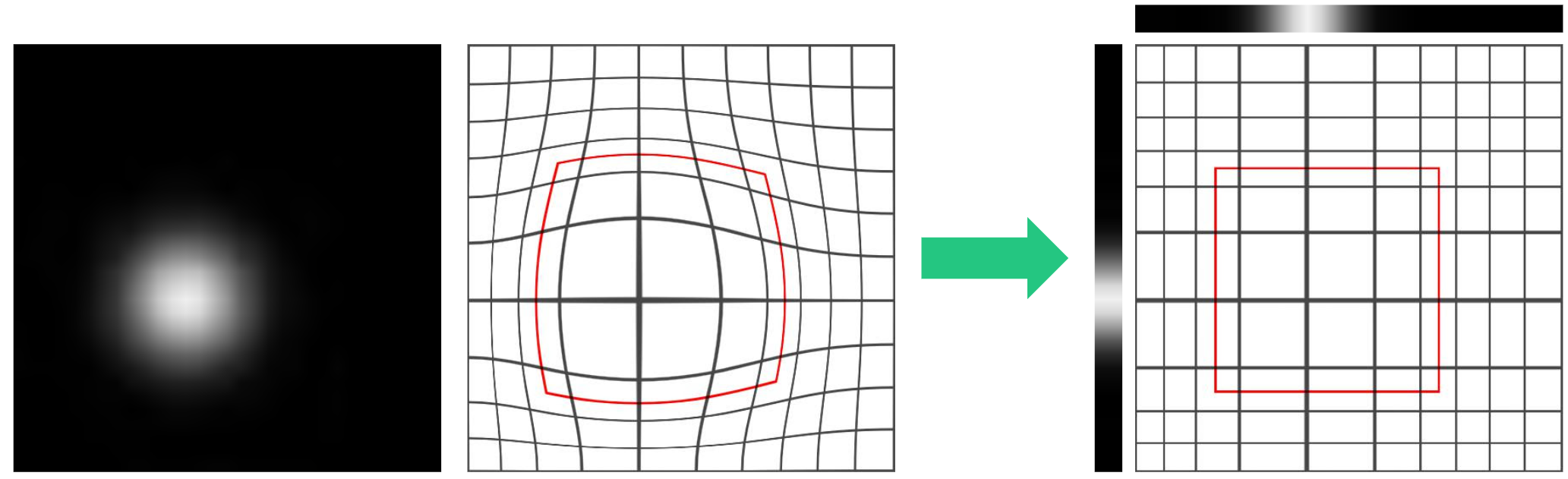}
\caption{By restricting the general class of warps ({\bf left})
to be separable ({\bf right}), we ensure that bounding boxes in the warped image (examples outlined in red) remain axis-aligned. We demonstrate that such regularization (surprisingly) improves performance, even though doing so theoretically restricts the range of expressible warps (details in Sec~\ref{direct saliency experiments}).
}
\label{fig:separable}
\end{figure}

\paragraph{Anti-Cropping Constraint}

We find the convolution form of saliency-guided spatial transform tends to crop the images, which might be acceptable for image classification where a large margin exists around the border. However, any cropping in object detection creates a chance to miss objects. We solve this by using reflect padding on the saliency map while applying the attraction kernel in Eq~\ref{torralba warp}. This introduces symmetries about each of the edges of the saliency map, eliminating all horizontal offsets along vertical image edges and vice versa. Thus cropping is impossible under this formulation.
A 1D illustration is shown in Fig \ref{fig:anticropping} to explain the problem and the solution.

\paragraph{Training formulation} Once we have the inference formulation, training is also straightforward as we require the loss $\loss$ to be computed in the original space: $\loss(\Q(f(\W_\T(I)), L)$, where $\Q$ is the label-type-specific backward mapping as shown in Fig~\ref{fig:warping}, and in our case, $\Q = \T^{-1}$. Note that $\W_\T$, $f$ and $\T^{-1}$ are all differentiable. While inference itself does not require the knowledge of $\T$, it is not the case for training detectors with region proposal networks (RPN) \cite{ren2015faster}.
When training RPNs \cite{ren2015faster}, the regression targets are the deltas between the anchors and the ground truth, and the deltas are later used in RoI Pooling/Align \cite{He2015SpatialPP,He2017MaskR}. The former should be computed in the original space (the ground truth is in the original space), while the latter is in the warped space (RoI Pooling/Align is over the warped image). This implies that the deltas need first to be learned in the original space, applied to the bounding box, and then mapped to the warped space using $\T$ for RoI Pooling/Align. But as discussed before, $\T$ cannot be easily computed. As a workaround,
we omit the delta encoding and adopt Generalized IoU (GIoU) loss \cite{rezatofighi2019generalized} to account for the lost stability.
The main idea of GIoU is to
better reflect the similarity of predicted and ground truth bounding boxes in cases of zero intersection; this has been shown to improve results.

\subsection{KDE Saliency Generator}
\label{Saliency Generator}
Prior work \cite{jaderberg2015spatial,recasens2018learning} trains a saliency network to generate saliency maps, which we explore as a baseline in Sec~\ref{sec:offline-exp}. Because 
saliency maps for object detection appear hard to learn,
we explore cheap alternatives for saliency map construction: dataset-level priors over object locations or temporal priors extracted from previous frame's predictions. Both priors can be operationalized with an approach that converts bounding boxes
to a saliency map.

\begin{figure}
\centering
\begin{subfigure}[h]{0.32\linewidth}
    \centering
    \includegraphics[width=1\linewidth]{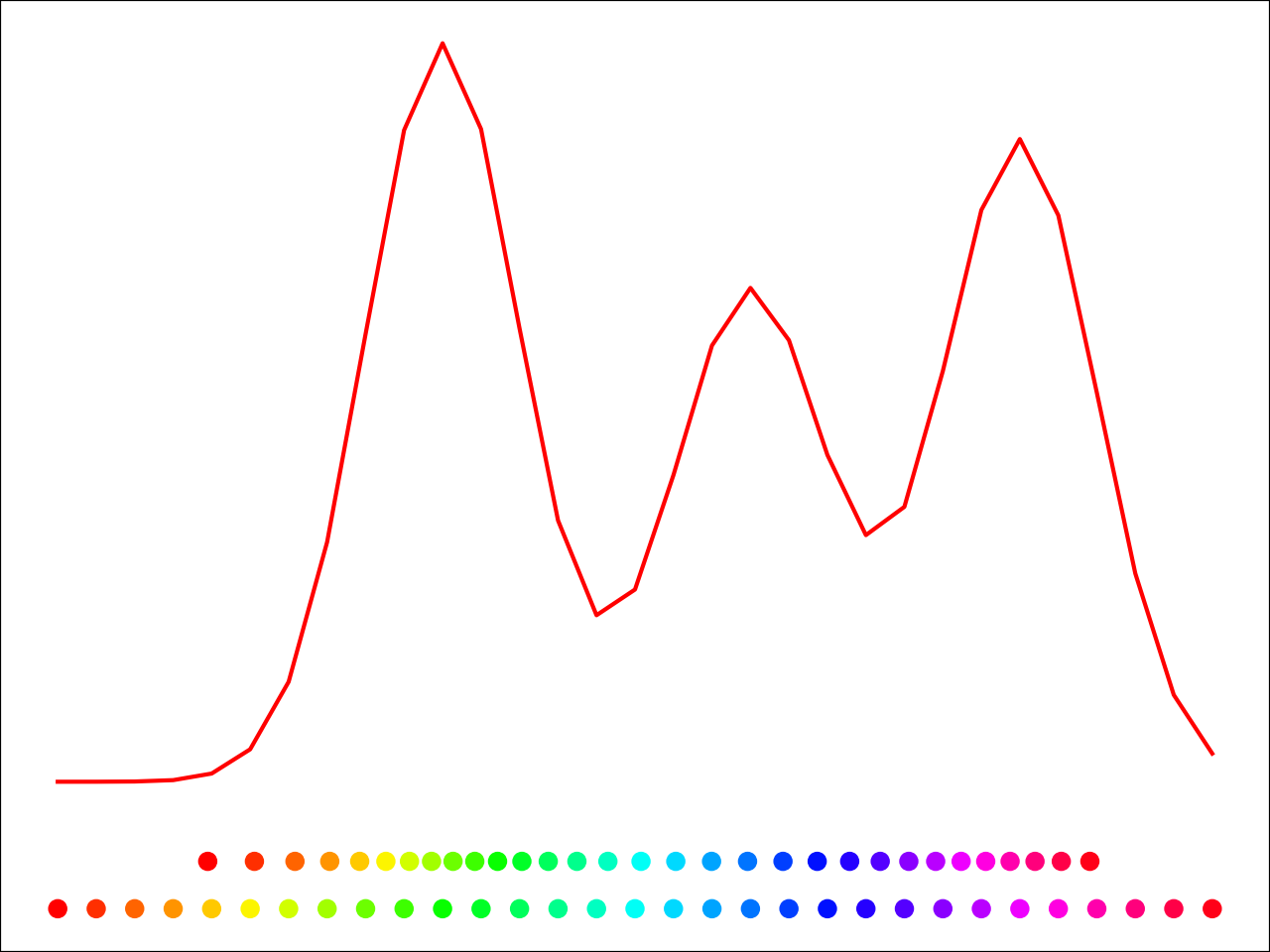}
    \subcaption{\scriptsize Default, $\sigma\approx 5.5$}
\end{subfigure}    
\begin{subfigure}[h]{0.32\linewidth}
    \centering
    \includegraphics[width=1\linewidth]{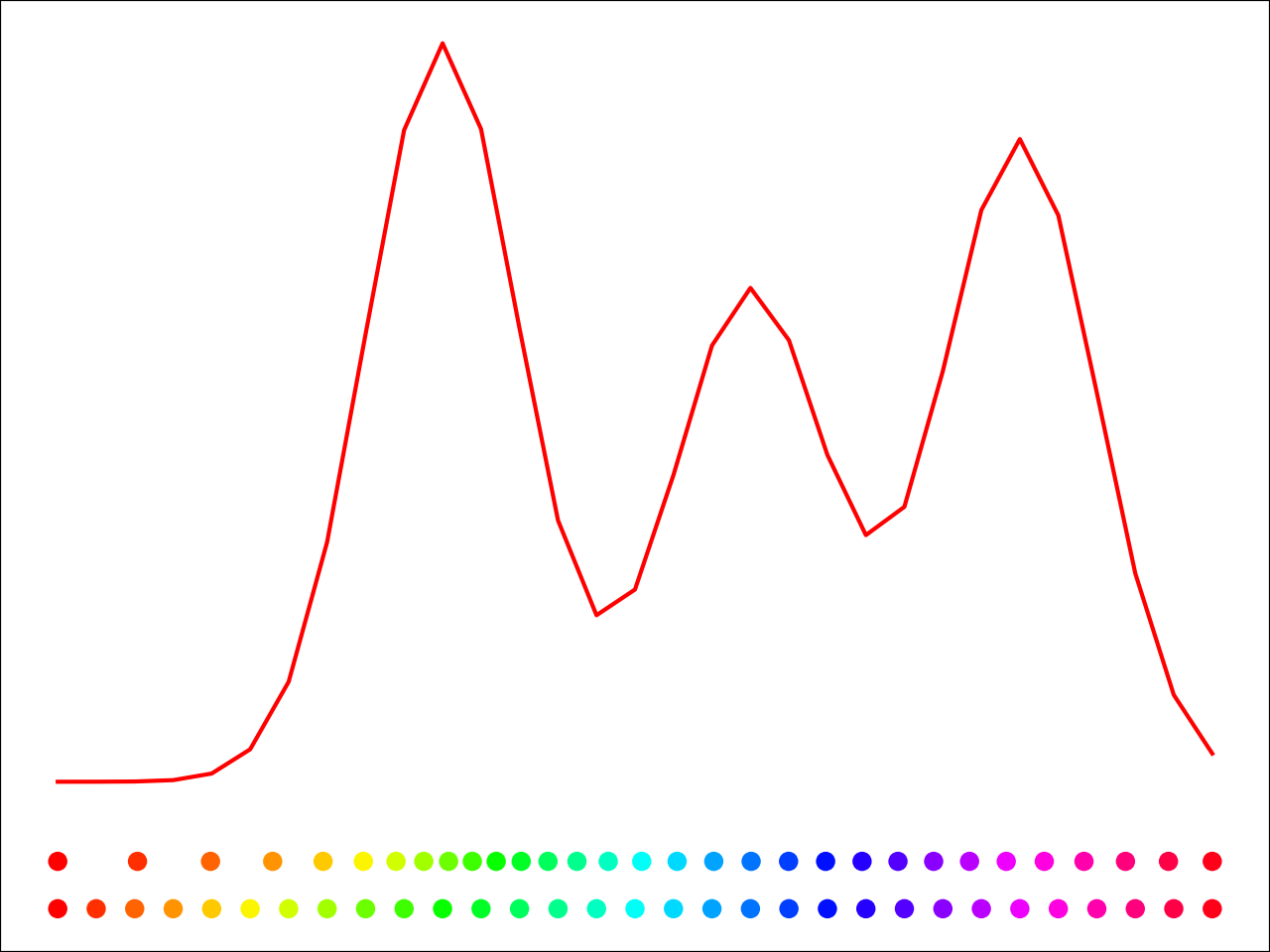}
    \scriptsize
    \subcaption{\scriptsize Anti-crop, $\sigma\approx 5.5$}
\end{subfigure}
\begin{subfigure}[h]{0.32\linewidth}
    \centering
    \includegraphics[width=1\linewidth]{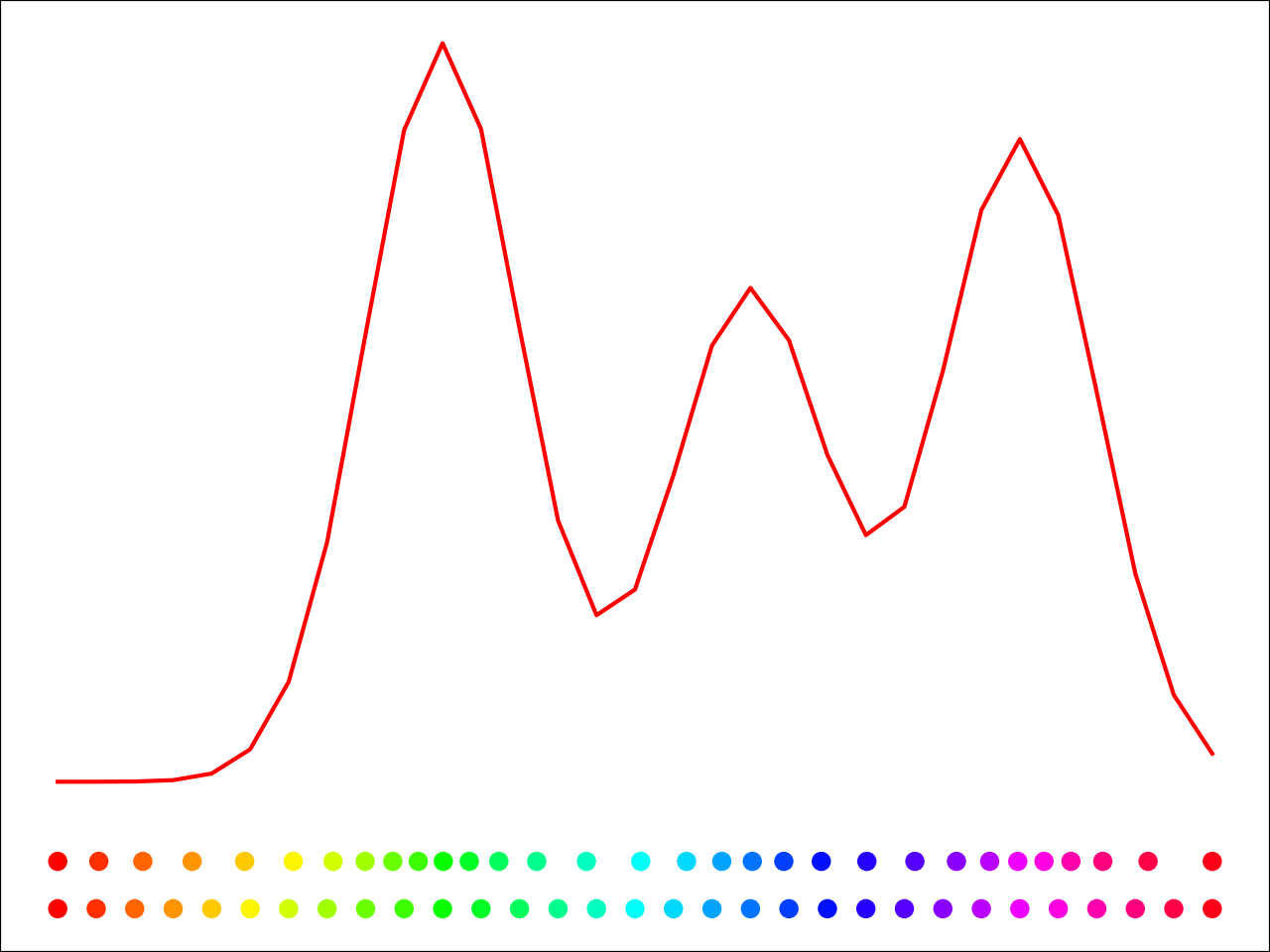}
    \scriptsize
    \subcaption{\scriptsize Anti-crop, $\sigma\approx 1.7$}
\end{subfigure}
\caption{Saliency-guided transform illustrated in 1D. The red curve is a saliency map $S$. The bottom row of dots are the output points (at uniform intervals), and the top row of dots are the locations where we've sampled each output point from the original ``image", as computed by applying $\T^{-1}_S$ to the output points. (a) The default transform can be understood as a weighted average over the output points and thus ignores points with near zero weights such as those at the boundaries. (b) Note the effects of introducing anti-crop reflect padding, and (c) how decreasing the std dev $\sigma$ of the attraction kernel $k$ results in more local warping around each peak (better for multimodal saliency distributions).}
\label{fig:anticropping}
\end{figure}

Intuitively, we build a saliency map by ``overlaying" boxes on top of one another via non-parametric kernel density estimation (KDE). More precisely, given a set of bounding boxes $B$ with centers $c_i$, heights $h_i$ and widths $w_i$, we model the saliency map $S_B$ as a sum of normal distributions:
\begin{align}
    S_B^{a, b} =
    \frac{1}{K^2} + a\sum_{(c_i, w_i, h_i) \in B} \mathcal N\left(c_i, b
    \begin{bmatrix}
    w_i & 0 \\
    0 & h_i
    \end{bmatrix}
    \right) \label{eq:kde}
\end{align}
where $a$ and $b$ are hyperparameters for amplitude and bandwidth, respectively, and $K$ is the size of the attraction kernel $k$ in Eq~\ref{torralba warp}. Adding the small constant is done to prevent extreme warps. We then normalize the 2D saliency map such that it sums to 1 and marginalize along the two axes if using the separable formulation\footnote{When using the separable formulation, we could instead skip the intermediate 2D saliency map representation. However, we opt not to, because the intermediate 2D saliency map produces more interpretable visualizations, and the difference in runtime is negligible.}.
As laid out in the previous section, this is then used to generate the image transformation $\mathcal T^{-1}_S$ according to Eq~\ref{torralba warp}. Ensuring that each kernel is locally normalized produces our desired behavior; we'll have high saliency for pixels covered by objects, and even higher saliency for pixels covered by small objects (that have their Gaussian mass focused on a smaller object size).

We can apply $S_B$ to the set of all bounding boxes in the training set to obtain a dataset-wide prior (denoted as $S_D$), or apply it to the previous frame's predictions to obtain a image-specific temporal prior (denoted as $S_I$). The former encodes dataset-level spatial priors such as small objects appearing near the horizon (Fig~\ref{fig:kde-visuals2}). The latter encodes a form of temporal contextual priming, allocating pixel samples to previously seen objects (with a default of uniform saliency for the first frame).
We also experiment with a weighted combination of both: $S_C = \alpha\cdot S_I + (1-\alpha)\cdot S_D$.  All of the above saliency generators are differentiable, so the final task loss can be used to learn hyperparameters $a,b,\alpha$. \deva{It sounds like we didn't backprop into $\alpha$. Would that help?}
\chittesh{It likely would, I hadn't considered that. I just implemented the learned version of S_I, because that had the best performance among the non-learned versions. I don't expect a huge improvement though.}

\section{Experiments}

We first compare FOVEA to naive downsampling on autonomous driving datasets such as Argoverse-HD. Next, we use streaming perception metrics to show that the accuracy gain is worth the additional cost in latency. Finally, we present results on BDD100K, showing the generalization of our method. We include
additional results, diagnostic experiments, and implementation details
in the 
appendix.

\subsection{Object Detection for Autonomous Navigation}
\label{sec:offline-exp}

Argoverse-HD \cite{Li2020StreamingP} is an object detection dataset for autonomous vehicles. Noteably, it contains high framerate (30 FPS) data and annotations.
As is standard practice, we adopt AP for evaluation.
We also report {\em end-to-end} latency (including image preprocessing, network inference, and bounding box postprocessing) measured on a single GTX 1080 Ti GPU. The image resolution for this dataset is $1920 \times 1200$, much larger than COCO's, which is capped at $640$. Since all models used in this paper are fully convolutional, we run them with different input scales, denoted by ratios to the native resolution, \eg, $0.5$x means an input resolution of $960 \times 600$.

\subsubsection{Baseline and Setup}
\label{Baseline and Setup}

The baseline we compare to throughout our experiments is Faster RCNN \cite{ren2015faster} with a ResNet-50 backbone \cite{he2016deep} plus FPN \cite{lin2017feature}. The default input scale for both the baseline and our method is $0.5$x.
For the baseline, however, we additionally train and test at $0.75$x and $1$x scales, to derive a sense of the latency-accuracy tradeoff using this model. Our contribution is orthogonal to the choice of the baseline detector and we obtain similar results with other detectors including RetinaNet \cite{lin2017focal} and YOLOF \cite{chen2021you} (shown in Appendix~\ref{app:other-det}). Additionally, we compare against other zoom-based approaches \cite{recasens2018learning,gao2018dynamic} in Appendix~\ref{app:other-baselines}.

Notably, Argoverse-HD's training set only contains {\em pseudo ground truth} (at the time of paper submission) generated by running high-performing detector HTC \cite{chen2019hybrid} in the offline setting. For all experiments, unless otherwise stated, we train on the train split with pseudo ground truth annotations, and evaluate on the val split with real annotations.
Additional measures are taken to prevent overfitting to biased annotations. We finetune COCO pretrained models
on Argoverse-HD for only $3$ epochs (\ie, early stopping). We use momentum SGD with a batch size of $8$, a learning rate of $0.02$, $0.9$ momentum, $10^{-4}$ weight decay, and a step-wise linear learning rate decay for this short schedule \cite{Li2020BudgetTrain}. Also, when training detectors with warped input, we apply our modifications to RPN and the loss function as discussed in Sec~\ref{sec:warping4det}.

\subsubsection{Learned Saliency}
\label{direct saliency experiments}

Our first control experiment does not make use of bounding box KDE priors, but rather directly learns a {\em global, dataset-wide} saliency map $S(x,y)$  via backprop.
We directly learn both separable and nonseparable saliency maps in Tab~\ref{final-table}. Training configuration and implementation details are given in Appendix~\ref{app:impl-details}. \deva{Would be cool to show these saliency maps in the appendix as well. For example, how visually similar is it to SD in Fig 7?}\chittesh{fair point}

\begin{figure}[t]
\centering
\includegraphics[width=0.49\linewidth]{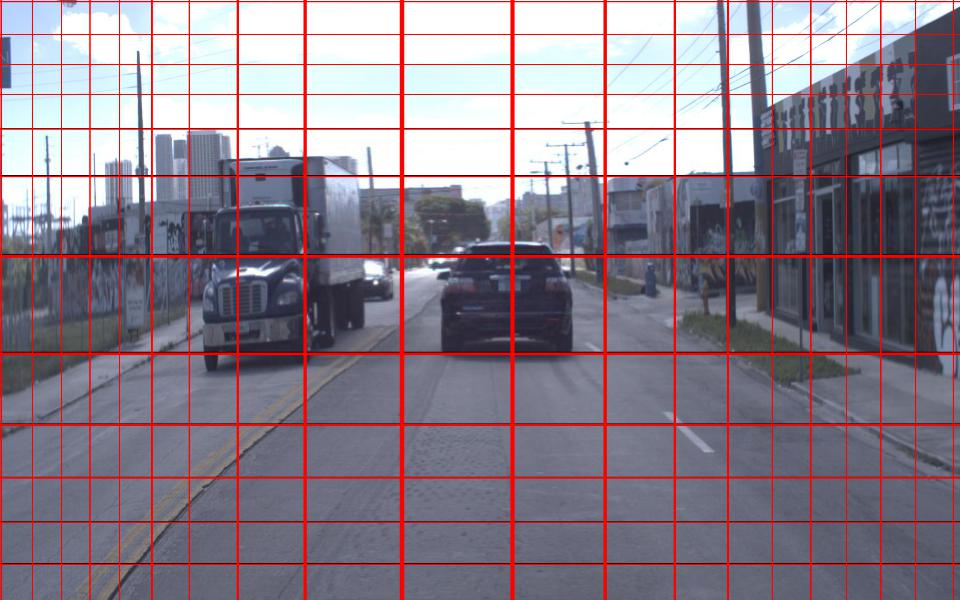}
\includegraphics[width=0.49\linewidth]{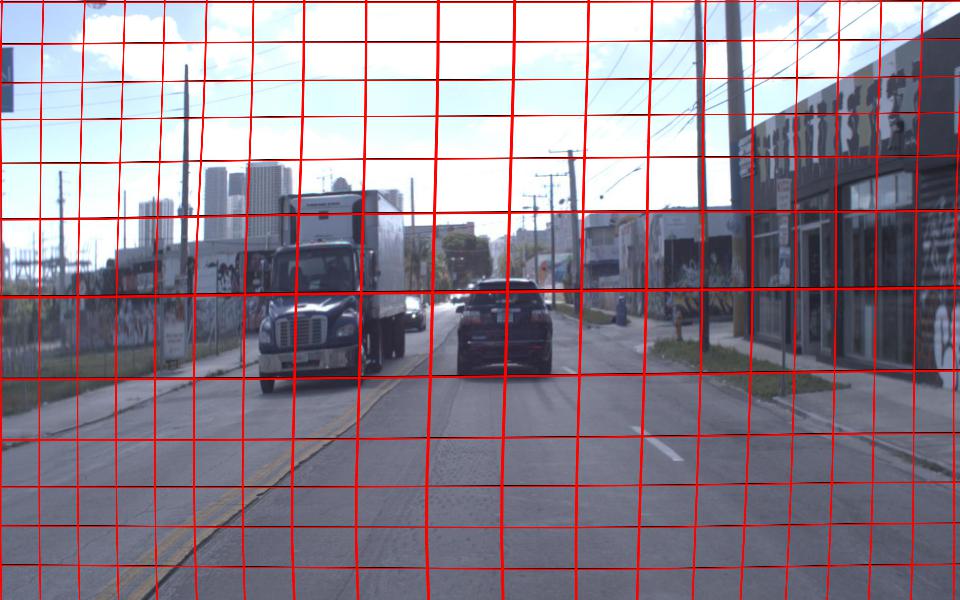}
\caption{The learned direct separable ({\bf left}) and nonseparable ({\bf right}) dataset-wide warps. Despite the vastly greater flexibility of nonseparable warps, the learned warp is almost separable anyway.}
\label{fig:direct-visuals}
\end{figure}

We find that both separable and nonseparable warps significantly improve overall AP over the baseline, owing to the boosted performance on small objects. However, there is also a small decrease in AP on large objects. Interestingly, even though nonseparable warps are more flexible, the learned solutions look nearly separable (Fig \ref{fig:direct-visuals}) but perform worse, indicating overfitting.
Therefore, going forward, we focus on separable warps in our experiments.

Following~\cite{recasens2018learning}, we also learn a ``saliency network" that maps each input image to its saliency map via a ResNet-18 backbone \cite{he2016deep}. In this sense, the learned saliency map would adapt to each image. However, we find that this approach very unstable for object detection. From our experiments, even with a small learning rate of $10^{-5}$ on the saliency network, the model learns a degeneracy in which an extreme warp leads to no proposals being matched with ground truth bounding boxes in the RoI bounding box head, leading to a regression loss of $0$.

\subsubsection{KDE Saliency Generator}

This section makes use of the KDE construction in Sec~\ref{Saliency Generator} to generate saliency maps. We first manually tune the amplitude $a$ and bandwidth $b$ to obtain desired magnifications.
We find that an amplitude $a=1$ and a bandwidth $b=64$ works the best, paired with an attraction kernel of std. dev. of about $17.8\%$ the image height, which allows for more local warps as illustrated in Fig \ref{fig:anticropping}. We finetune our models using the same configuration as the baseline, the only difference being the added bounding box and saliency-guided spatial transformation layer. We learn $S_D$ using all bounding boxes from the training set and for simplicity, learn $S_I$ with jittered ground-truth boxes from the current frame (though at test-time it always uses predictions from the previous frame). We set $\alpha=0.5$ for $S_C$.

We then learn hyperparameters $a$ and $b$ through backpropagation, since our KDE formulation is differentiable. We initialize parameters $a'$ and $b'$ to 0, under the construction that $a=|1+a'|+0.1, b=64\cdot |1+b'|+0.1$.
The learning rate of $a'$ and $b'$ is set to $10^{-4}$ with zero weight decay. Other than this, we train the learned KDE (LKDE) model with the same configuration as the baseline. We implement the $S_I$ formulation.

All results are shown in Table~\ref{final-table}. Even without finetuning our detector, using a simple fixed dataset-wide warp $S_D$, we find significant improvements in AP. As we migrate to temporal priors with finetuning, we see even more improvement. As in the learned saliency case, these improvements in overall AP are due to large boosts in AP$_S$, outweighing the small decreases in AP$_L$. Combining our saliency signals ($S_C$) doesn't help, because in our case, it seems that the temporal signal is strictly stronger than the dataset-wide signal. Perhaps if we had an alternate source of saliency like a map overlay, combining saliencies could help. Our best method overall is LKDE, which learned optimal values $a=1.07, b=71.6$.
Learning a nonseparable saliency performs better than our hand-constructed dataset-wide warp $S_D$; however, they're both outperformed by $S_I$.
Importantly, our LKDE not only significantly improves AP$_S$, but also improves {\em all} other accuracy measures, suggesting that our method does not need to tradeoff accuracy of large objects for that of small objects.
Finally, we note that our increased performance comes at the cost of only about $2$ ms in latency.
\deva{Could you try SI where the first-frame uniform saliency map is replaced by SD? The numbers suggest this might work the best. You could also try learning $\alpha$, but not sure this would help.} \chittesh{Honestly I feel like these would lead to negligible improvements.}

\begin{figure}[t]
\centering
\includegraphics[width=0.9\linewidth]{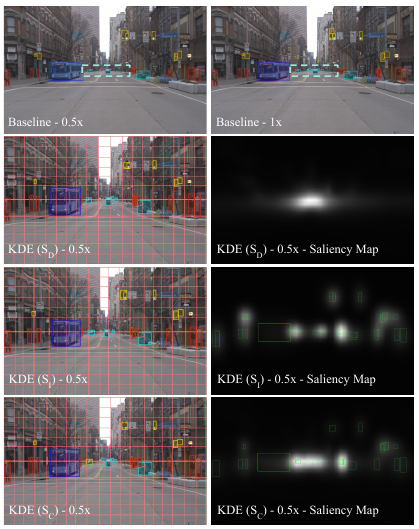}
\caption{Qualitative results for our methods after finetuning on Argoverse-HD. The cars in the distance (in the dotted boxes), undetected at $0.5$x scale, are detected at $1$x scale, and partially detected by our methods. Different rows show the variations within our method based on the source of attention.
} 
\label{fig:kde-visuals2}
\end{figure}

\begin{table*}[ht]
    \centering
    Argoverse-HD before finetuning\\
    \begin{adjustbox}{width=1.0\linewidth,center}
    \begin{tabular}{@{}lccccccccccccccc@{}}
    \toprule
    Method    & AP & AP$_{50}$ & AP$_{75}$ & AP$_{S}$ & AP$_{M}$ & AP$_{L}$ & person & mbike & tffclight & bike & bus & stop & car & truck & Latency (ms) \\
    \midrule
    Baseline & 21.5	&35.8	&22.3	&2.8	&22.4	&\textbf{50.6}	&20.8	&9.1	&13.9	&7.1	&48.0	&16.1	&37.2	&20.2& 49.4 $\pm$ 1.0 \\
    \midrule
    KDE ($S_D$) & 23.3	&40.0	&22.9	&5.4	&25.5	&48.9	&20.9	&13.7&	12.2	&9.3&	\textbf{50.6}&	20.1&	40.0&	19.5&	52.0 $\pm$ 1.0 \\
    KDE ($S_I$)  &\textbf{24.1}	&\textbf{40.7}	&\textbf{24.3}	&\textbf{8.5}	&24.5	&48.3	&\textbf{23.0}	&\textbf{17.7}	&\textbf{15.1}	&\textbf{10.0}	&49.5	&17.5	&\textbf{41.0}	&19.4& 51.2 $\pm$ 0.7\\
    KDE ($S_C$) & 24.0	&40.5&	\textbf{24.3}&	7.4&	\textbf{26.0}&	48.2	&22.5	&14.9	&14.0	&9.5	&49.7	&\textbf{20.6}	&\textbf{41.0}	&\textbf{19.9}	&52.0 $\pm$ 1.2 \\
    \midrule
    Upp. Bound ($0.75$x)  & 27.6	&45.1	&28.2	&7.9	&30.8	&51.9	&29.7	&14.3	&21.5	&6.6	&54.4	&25.6	&44.7	&23.7& 86.9 $\pm$ 1.6\\
    Upp. Bound ($1.0$x) &32.7	&51.9	&34.3	&14.4	&35.6	&51.8	&33.7	&21.1	&33.1	&5.7	&57.2	&36.7	&49.5	&24.6& 133.9 $\pm$ 2.2	\\
    \bottomrule
    \end{tabular}
    \end{adjustbox}
    Argoverse-HD after finetuning\\
    \begin{adjustbox}{width=1.0\linewidth,center}
    \begin{tabular}{@{}lccccccccccccccc@{}}
    \toprule
    Method    & AP & AP$_{50}$ & AP$_{75}$ & AP$_{S}$ & AP$_{M}$ & AP$_{L}$ & person & mbike & tffclight & bike & bus & stop & car & truck & Latency (ms) \\
    \midrule
    Baseline &24.2	&38.9	&26.1	&4.9	&29.0	&50.9	&22.8	&7.5	&23.3	&5.9	&44.6	&19.3	&43.7	&26.6 & 50.9 $\pm$ 0.9\\
    \midrule
    Learned Sep. & 27.2&44.8&28.3&\textbf{12.2}&29.1&46.6&24.2&	14.0	&22.6&	7.7	&39.5&	\textbf{31.8}&	50.0&	27.8&51.5 $\pm$ 1.0\\
    Learned Nonsep. & 25.9&42.9&26.5&10.0&28.4&48.5&25.2	&11.9	&20.9&	7.1&	39.5&	25.1&	49.4&	28.1&50.0 $\pm$ 0.8\\
    \midrule
    KDE ($S_D$) & 26.7	&43.3&	27.8&	8.2	&29.7&	54.1	&25.4&	13.5&	22.0	&8.0	&\textbf{45.9}&	21.3&	48.1	&29.3&	50.8 $\pm$ 1.2\\
    KDE ($S_I$) & 28.0	& 45.5	&\textbf{29.2}	&10.4	&\textbf{31.0}	&\textbf{54.5}	&27.3	&16.9	&\textbf{24.3}	&\textbf{9.0}	&44.5	&23.2	&\textbf{50.5}	&28.4 & 52.2 $\pm$ 0.9 \\
    KDE ($S_C$) &27.2&	44.7&	28.4	&9.1	&30.9&	53.6	&27.4	&14.5&	23.0&	7.0	&44.8	&21.9&	49.9&	\textbf{29.5}	&52.1 $\pm$ 0.9 \\
    \midrule
    LKDE ($S_I$) & \textbf{28.1}	&\textbf{45.9}	&28.9	&10.3&	30.9	&54.1	&\textbf{27.5}&	\textbf{17.9}&	23.6&	8.1	&45.4	&23.1&	50.2&	28.7	&50.5 $\pm$ 0.8 \\
    \midrule
    Upp. Bound ($0.75$x) & 29.2	&47.6	&31.1	&11.6	&32.1	&53.3	&29.6	&12.7	&30.8	&7.9	&44.1	&29.8	&48.8	&30.1& 87.0 $\pm$ 1.4 \\
    Upper Bound ($1.0$x) & 33.3 & 53.9 & 35.0 & 16.8 & 34.8 & 53.6 & 33.1 & 20.9 & 38.7 & 6.7 & 44.7 & 36.7 & 52.7 & 32.7 & 135.0 $\pm$ 1.6 \\
    \bottomrule
    \end{tabular}
    \end{adjustbox}
    \caption{Results before and after finetuning on Argoverse-HD. Without retraining, processing warped images (KDE $S_I$, top table) improves overall AP by 2.6 points and triples AP$_S$.
    Even larger gains can be observed after finetuning, making our final solution (LKDE $S_I$) performing close to the $0.75$x upper bound.
    Please refer to the text for a more detailed discussion.}
    \label{final-table}
\end{table*}

\subsection{Streaming Accuracy for Cost-Performance Evaluation}
\label{sec:streaming}

{\em Streaming accuracy} is a metric that coherently integrates latency into standard accuracy evaluation and therefore is able to {\em quantitatively measure the accuracy-latency tradeoff} for embodied perception \cite{Li2020StreamingP}.
\ifunlimited
Such a setup is achieved by having the benchmark stream the data to the algorithm in real-time and query for the state of the world at all time instants. One of their key observations is that by the algorithm finishes processing, the world has around changed and therefore proper temporal scheduling and forecasting methods should be used to compensate for this latency.
\else
\fi
Here we adopt their evaluation protocol for our cost-performance analysis. In our case of streaming object detection, the streaming accuracy refers to {\em streaming AP}. We use the same GPU (GTX 1080 Ti) and their public available codebase for a fair comparison with their proposed solution. Their proposed solution includes a scale-tuned detector (Faster R-CNN), dynamic scheduler (shrinking-tail) and Kalman Filter forecastor. Our experiments focus on improving the detector and we keep the scheduler and forecastor fixed.

Tab~\ref{tab:streaming-full} presents our evaluation under the full-stack setting (a table for the detection-only setting is included in Appendix~\ref{app:det-streaming}.
We see that FOVEA greatly improves the previous state-of-the-art. The improvement first comes from a faster and slightly more accurate implementation of the baseline (please refer to Appendix~\ref{app:impl-details} for the implementation details). Note that under streaming perception, a faster algorithm while maintaining the same offline accuracy translates to an algorithm with higher streaming accuracy. The second improvement is due to training on pseudo ground truth (discussed in Sec~\ref{Baseline and Setup}). Importantly, our KDE image warping further boosts the streaming accuracy significantly {\em on top of} these improvements. Overall, these results suggest that {\em image warping is a cost-efficient way to improve accuracy}.

\begin{table}[t]
\small
\centering
\begin{tabular}{clcccc}
\toprule
ID & Method                                             & AP            & AP$_S$       & AP$_M$        & AP$_L$        \\
\midrule
1  & Prior art \cite{Li2020StreamingP} & 17.8          & 3.2          & 16.3          & 33.3          \\
\midrule
2  & + Better implementation                            & 19.3          & 4.1          & 18.3          & 34.9          \\
3  & + Train with pseudo GT                          & 21.2          & 3.7          & \textbf{23.9}          & 43.8          \\
\midrule
4  & 2 + Ours ($S_I$)                                     & 19.3          & 5.2          & 18.5          & 39.0          \\
5  & 3 + Ours ($S_I$)                                     & \textbf{23.0} & \textbf{7.0} & 23.7 & \textbf{44.9}           \\
\bottomrule
\end{tabular}
\caption{Streaming evaluation in the full-stack (with forecasting) setting on Argoverse-HD. We show that our proposed method significantly improves previous state-of-the-art by 5.2, in which 1.5 is from better implementation, 1.9 is from making use of pseudo ground truth and 1.8 is from our proposed KDE warping.}
\label{tab:streaming-full}
\end{table}

\subsection{Cross-Dataset Generalization}

\begin{table}[]
\small
\centering
\begin{tabular}{clcccccc}
\toprule
ID & Method             & AP            & AP$_S$       & AP$_M$        & AP$_L$        \\
\midrule
1  & Baseline ($0.5$x)    & 15.1          & 1.0          & 10.6          & \textbf{39.0} \\
2  & Ours $S_D$ ($0.5$x)  & 13.7          & 1.3          & 10.0          & 34.7          \\
3  & Ours $S_I$ ($0.5$x)  & \textbf{16.4} & \textbf{2.1} & \textbf{12.8} & 38.6          \\
\midrule
4  & Baseline ($0.75$x)   & 19.7          & 3.0          & 16.1          & \textbf{44.2} \\
5  & Ours $S_D$ ($0.75$x) & 18.2          & 3.4          & 15.4          & 40.0          \\
6  & Ours $S_I$ ($0.75$x) & \textbf{20.1} & \textbf{5.2} & \textbf{17.0} & 42.5          \\
\midrule
7  & Upper bound ($1.0$x) & 22.6          & 5.7          & 20.1          & 45.7    \\
\bottomrule
\end{tabular}
\caption{Cross-dataset generalization to BDD100K \cite{bdd100k}. Rows 2 \& 5 are saliency computed on the Argoverse-HD training set, as expected, they fail to generalize to a novel dataset. Despite operating at a larger temporal stride (5 FPS vs 30 FPS), our proposed image-adaptive KDE warping generalizes to a novel dataset (row 3 \& 6). Note that here the image native resolution is smaller at $1280 \times 720$.}
\label{tab:bdd}
\end{table}

Our experiments so far are all conducted on the Argoverse-HD dataset. In this section, we cross-validate our proposed method on another autonomous driving dataset BDD100K \cite{bdd100k}. Note that BDD100K and Argoverse-HD are collected in different cities. For simplicity, we only test out {\em off-the-shelf} generalization without any finetuning. We experiment on the validation split of the MOT2020 subset, which contains 200 videos with 2D bounding boxes annotated at 5 FPS (40K frames in total). Also, we only evaluate on common classes between BDD100K and Argoverse-HD: person, bicycle, car, motorcycle, bus, and truck. The results are summarized in Tab~\ref{tab:bdd}, which demonstrate the generalization capability of our proposed method.

\section{Conclusion}

We propose FOVEA, a highly efficient attentional model for object detection. Our model magnifies regions likely to contain objects, making use of top-down saliency priors learned from a dataset or from temporal context.
To do so, we make use of differentiable image warping that ensures bounding box predictions can be mapped back to the original image space.
The proposed approach significantly improves over the baselines on Argoverse-HD and BDD100K.
For future work, it would be natural to make use of trajectory forecasting models to provide even more accurate saliency maps for online processing.
    \bigskip
    
\noindent {\bf Acknowledgements:} This work was supported by the CMU Argo AI Center for Autonomous Vehicle Research.

\fi

{\small
\bibliographystyle{ieee_fullname}
\bibliography{egbib}
}

\ifstandalonesupplement
\else
    \ifappendix
        \clearpage
        \renewcommand{\thesection}{\Alph{section}}
        \renewcommand{\thefigure}{\Alph{figure}}
        \renewcommand{\thetable}{\Alph{table}}
        \setcounter{section}{0}
        \setcounter{figure}{0}
        \setcounter{table}{0}
        \renewcommand*{\theHsection}{A\the\value{section}}
        \renewcommand*{\theHfigure}{A\the\value{figure}}
        \renewcommand*{\theHtable}{A\the\value{table}}
        
    \fi
\fi


\end{document}